%% file: DiffAuction.tex
\documentclass{article} 
\usepackage{manuscript,times}

\input{math_commands.tex}

\usepackage{hyperref}
\usepackage{url}
\usepackage{longtable}
\usepackage{array}
\newcolumntype{L}[1]{>{\raggedright\arraybackslash}p{#1}}
\newcolumntype{C}[1]{>{\centering\arraybackslash}p{#1}}
\usepackage{graphicx} 
\usepackage{booktabs} 
\usepackage{float}    
\usepackage{capt-of}
\usepackage{makecell}
\usepackage{xcolor}
\usepackage{subcaption}
\usepackage{listings}
\lstdefinestyle{promptstyle}{
    basicstyle=\ttfamily\footnotesize,
    breaklines=true,
    breakatwhitespace=false,
    columns=fullflexible,
    keepspaces=true,
    showstringspaces=false,
    frame=single,
    rulecolor=\color{black!30},
    xleftmargin=0.5em,
    xrightmargin=0.5em,
    framexleftmargin=0.4em,
    framexrightmargin=0.4em,
    aboveskip=0.8em,
    belowskip=0.8em
}

\lstnewenvironment{promptbox}[1]
{
    \lstset{
        style=promptstyle,
        title={#1},
        captionpos=t
    }
}
{}
\usepackage{listings}
\renewcommand{\arraystretch}{1.08}

\title{DiffAuction: An Auction-Driven Framework for Generative Image Advertising}

\author{Yijie Su\textsuperscript{1,$\dagger$}, Ruiyang Si\textsuperscript{$\dagger$}, Tonghan Wang\textsuperscript{2,3,$\diamond$}\\
\normalfont\textsuperscript{1}Tongji University\quad\textsuperscript{2}Tsinghua University\quad\textsuperscript{3}Shanghai Qi Zhi Institute\\
\normalfont\textsuperscript{$\diamond$}Correspondence to: \texttt{thw@mail.tsinghua.edu.cn}}

\begin{document}

\maketitle
\begingroup
\renewcommand{\thefootnote}{\fnsymbol{footnote}}
\footnotetext[2]{Work done during an internship at Tsinghua University.}
\endgroup

\begin{abstract}
Diffusion models create an opportunity for advertising in generated visual content, with multiple advertisers competing for exposure within a shared image. Yet existing approaches are computationally costly when generating images under varying target-share configurations, and it remains unclear how the resulting outcomes should be evaluated and priced. To address these gaps, we propose DiffAuction, an auction framework connecting controllable generation, click-intent evaluation, and auction design. First, our training-free Residual Share Calibration (RSC) decomposes joint-attention updates into product-specific residuals and rescales them according to target shares. Because a fixed prompt tends to produce similar viewpoints that favor the same products regardless of their target shares, we further introduce Share-Conditioned Prompting (SCP) to adapt the viewpoint to each desired allocation. Second, we develop an evaluator that uses a VLM to extract visual evidence and an LLM to assess Click-Through Intent (CTI), a proxy for click propensity, along four dimensions: Product Prominence, Brand Recognizability, Scene Integration, and Contextual Value Communication. We validate its assessments against judgments from 10 trained annotators on 1,000 images. Third, we formulate this advertising setting as an auction over jointly feasible generation configurations. Each configuration has a public expected CTI vector, while advertisers hold configuration-specific private values per unit of CTI that map click intent into commercial value. We adopt a learnable affine maximizer auction (AMA), which jointly selects a shared generation configuration and determines advertiser-specific payments to maximize expected platform revenue. The resulting allocation and payment rules guarantee dominant-strategy incentive compatibility (DSIC) and individual rationality (IR). Experiments show that our generation method strengthens the monotonic relationship between target shares and CTI relative to baselines, while the learned AMA achieves higher platform revenue in the resulting auction setting.

\end{abstract}

\section{Introduction}
\label{sec:introduction}
Online advertising monetizes user attention by allocating exposure through auctions \citep{edelman2007internet,varian2007position}. Whereas conventional auctions assign predefined ad slots, a growing body of work on LLM-native ad has begun to auction influence over generated content based on advertisers’ bids and preferences \citep{duetting2024mechanism,hajiaghayi2024ad,soumalias2026truthful,zhao2025llm,yun2026llm,zhang2026naiad,zhang2026pila,liu2026incentive,turner2026evaluating}. These advances demonstrate the potential of treating generated content itself as advertising inventory.

Diffusion models extend this opportunity to visual advertising, raising the possibility of advertisers competing for exposure within a shared generated image. Compared with LLM-based ad auctions, this direction remains underexplored. Designing such auctions requires addressing three interdependent challenges.
First, the generation process should realize the advertiser exposure shares selected by the auction, yet diffusion models offer limited control over these shares through prompting alone.
Second, the effectiveness of each advertiser's exposure must be measurable. Click-through intent (CTI) provides a suitable criterion. Whereas conventional auctions can draw on historical click-through rates for pre-defined slots, each generated image is a new outcome whose product-level CTI must be estimated from its content.
Third, the mechanism should combine the publicly known expected CTI with advertisers' private
values per CTI unit, select a jointly feasible exposure-share configuration, and
determine bidder-specific payments that incentivize truthful reporting.

To address the first challenge, \citet{sun2025bidding} introduce
bid-weighted score composition, combining multiple diffusion predictions, each
conditioned on a distinct subset of product prompts. This requires multiple conditional model evaluations per timestep, increasing
computational cost as the number of products grows. To accelerate generation, we condition the model jointly on all product prompts and incorporate the exposure-share specification within a single forward pass per timestep.
We implement this using Residual Share Calibration (RSC), a training-free method that decomposes joint attention's residual output into product-specific updates induced by the corresponding product tokens.
These updates capture how each product shifts the latent image features during generation. RSC calibrates their magnitudes according to the target exposure shares.
A fixed prompt can favor similar viewpoints even as target shares change, leaving certain products consistently obscured or overly prominent. We therefore introduce Share-Conditioned Prompting (SCP) that adjusts the viewing perspective according to target shares to better support the desired exposure allocation. 

To address the second challenge, we introduce a product-level CTI proxy based on four interpretable dimensions: product prominence, brand recognizability, scene integration, and contextual value communication, with an analysis supporting their coverage of the principal aspects of CTI evaluation. We use a VLM to extract visual evidence for each product, and an LLM to score the evidence along each dimension using a fixed rubric. The aggregate score provides a content-based estimate of product-level
click intent, which we use to evaluate the effectiveness of the realized
exposure.
 We assess this evaluator on a dataset of 1,000 images annotated by ten annotators and find that its scores correlate strongly with human ratings.  

To address the third challenge, we formulate exposure trading in generative image advertising as an auction over generation configurations subject to joint feasibility constraints. For a given scene and set of competing advertisers, the platform offers candidate generation configurations, each specifying an exposure share and an expected CTI for each advertiser. 
Moreover, different ad presentations may affect both click intent and the conversion potential of the resulting attention \citep{schwartz2017customer,li2025impact}. Advertisers may thus value each unit of CTI differently across configurations based on their private conversion data and campaign objectives.
We therefore allow each advertiser to hold multiple private valuations, one for each configuration.
Building on this formulation, we adopt a learnable affine maximizer auction (AMA)~\citep{curry2023differentiable} to optimize expected platform revenue while ensuring truthful reporting. Using a joint scoring function with learnable advertiser weights and configuration-specific offsets, the mechanism selects the highest-scoring configuration and constructs advertiser-specific price menus that depend only on the other advertisers’ reports. We show that this allocation and payment rule make the selected configuration utility-maximizing for each advertiser within its own menu,
and that this mechanism guarantees dominant-strategy incentive compatibility (DSIC) and individual rationality (IR). 

Our experiments show that combining RSC with SCP reduces generation time while strengthening the monotonic relationship between target exposure shares and product-level CTI. Our CTI evaluation also exhibits relatively high agreement with human annotations. Finally, experiments with simulated private valuations demonstrate that the learned AMA achieves higher platform revenue than the baselines in this auction setting.

In summary, our contributions are threefold:
\begin{itemize}
    \item We propose a training-free method combining RSC and SCP to control product exposure in generated images according to target shares.
    \item We develop a VLM--LLM evaluator that assesses product-level CTI along four interpretable dimensions and construct a dataset of 1,000 images annotated by 10 annotators to evaluate its agreement with human judgments.
    \item We formulate generative image advertising as an auction over
jointly feasible generation configurations with multiple
configuration-specific private values per CTI unit. We instantiate
this formulation with a learnable affine maximizer auction (AMA) that
optimizes expected platform revenue while guaranteeing DSIC and IR.
\end{itemize}

\newpage
\section{Related Work}
\label{sec:related works}

\paragraph{Auction-based generative advertising.}
Generative advertising treats model-generated content as advertising inventory, allowing advertisers to bid for influence beyond predefined ad slots. Existing LLM-native mechanisms realize this idea at different levels: token-level auctions allow bids to influence individual tokens \citep{duetting2024mechanism}; RAG-based segment auctions allocate ads to discourse segments according to bids and relevance \citep{hajiaghayi2024ad}; MOSAIC aggregates advertisers’ preferences over generated responses without fine-tuning the LLM \citep{soumalias2026truthful}; and LLM-Auction jointly models auction allocation and response generation \citep{zhao2025llm}. Neuron Auctions shift the auction object from surface text to internal FFN representations and treat continuous intervention budgets as auctionable commodities \citep{yun2026llm}. These studies motivate our formulation of product exposure in a generated image as an auctionable resource.

However, auction-driven generative image advertising remains relatively underexplored. \citet{sun2025bidding} provide an initial study. Their target-share-conditioned generation requires multiple conditional model evaluations per denoising timestep, causing computational cost to grow with the number of products. Moreover, they use CLIP similarity as a proxy for product value, but this simple proxy lacks interpretability and direct validation. They also do not consider the more complex setting in which advertisers hold multiple configuration-specific private valuations, or integrate image generation and auction design into a unified framework for optimizing platform revenue. Therefore, systematic work is still needed to fill these research gaps.



\paragraph{Training-free controllable image generation.}
A broad line of research controls pretrained diffusion models at inference time without additional model training. Existing approaches intervene at different stages of the generation process. Attention-based methods manipulate token--image interactions to preserve, amplify, or suppress specific concepts; for example, Prompt-to-Prompt edits cross-attention maps and supports token-level attention reweighting~\citep{hertz2022prompt}, while Rich Text introduces explicit token and region-aware guidance~\citep{ge2023expressive}. Guidance-based approaches modify denoising directions to strengthen or suppress semantic concepts, as exemplified by SEGA~\citep{brack2023sega}. More recent methods directly edit internal representations of diffusion Transformers; FluxSpace, for instance, identifies editable directions in FLUX feature spaces~\citep{dalva2025fluxspace}. These methods demonstrate the effectiveness of inference-time intervention for semantic and compositional control. However, our setting differs from conventional concept editing: multiple advertised products coexist in the same jointly generated scene, and the objective is not merely to include or emphasize a concept, but to 
regulate their relative exposure according to a prescribed share vector. 
\citet{sun2025bidding} address this problem by combining generation directions obtained from multiple conditional model evaluations, resulting in increased computational cost as the number of products grows. In contrast, the combination of Share-Conditioned Prompting (SCP) and Residual Share Calibration (RSC) supports target-share-conditioned generation with only a single model forward pass per denoising timestep.

\section{Method}

\begin{figure}[t]
    \centering
    \includegraphics[width=1.0\linewidth]{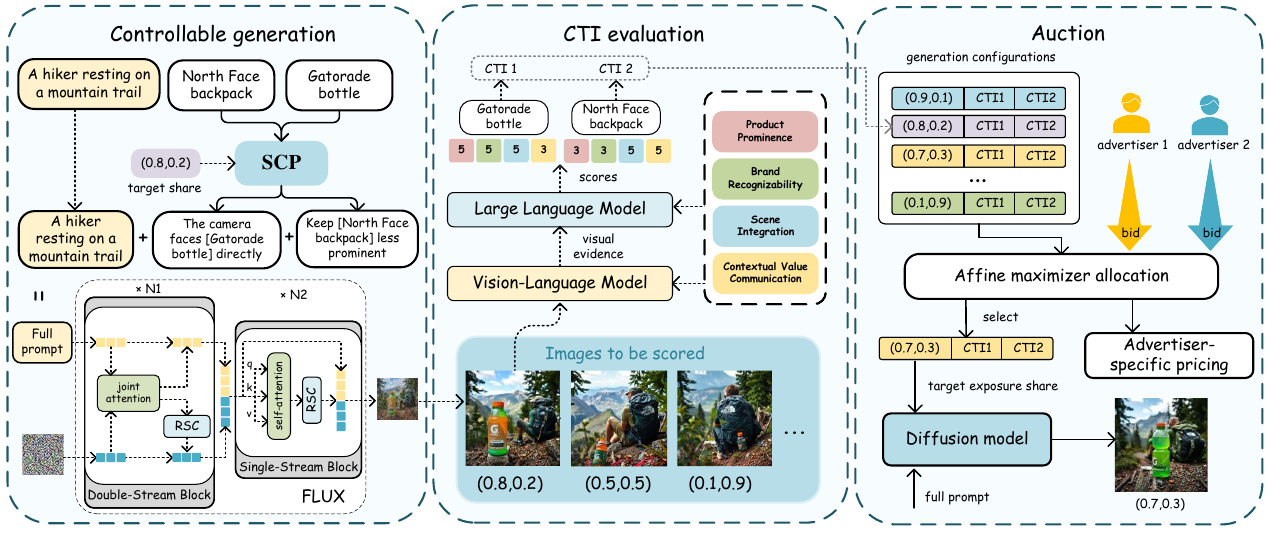}
    \caption{Overview of DiffAuction. DiffAuction connects generation, CTI evaluation, and auction. Given target shares, RSC and SCP control the exposure of products during generation. The resulting images are evaluated offline by a VLM--LLM pipeline
that estimates product-level CTI. During the auction, a
learnable affine maximizer combines the expected CTI of each
configuration with advertisers' reported values per CTI unit to
select a shared configuration and determine advertiser-specific
payments. The selected exposure shares are used to generate the advertisement image.  }
    \label{fig:overview}
\end{figure}

As illustrated in Figure~\ref{fig:overview}, target shares, serving as control inputs to the generative model, determine the intended exposure allocation; CTI estimates the product-level click intent associated with the
resulting exposure, and the auction combines the expected CTI of each
candidate configuration with advertisers' reported values per CTI unit
for allocation and pricing. We describe each component in the following sections.

\subsection{Residual Share Calibration and Share-Conditioned Prompting}
\label{sec:rsc}

Given a scene prompt and $N$ product prompts, we use a target-share vector
$\mathbf{s}=(s_1,\ldots,s_N)\in\Delta^{N-1}$ to condition the relative
exposure of the products. We propose Residual Share Calibration (RSC),
a training-free method that decomposes the image-feature update from joint
attention into product-specific residuals and rescales them so that their
shares of the summed product-wise Frobenius norms match the target vector.

\paragraph{Product-wise residual decomposition.}
Consider one joint-attention block and omit its denoising-step and block
indices for clarity. Let $\mathbf{A}\in\R^{H\times I\times T}$ denote the
image-to-text attention tensor and $\mathbf{V}\in\R^{H\times T\times d_h}$
the corresponding text-value tensor. Here, $H$ is the number of attention
heads, $I$ is the number of image tokens, $T$ is the number of text tokens,
and $d_h$ is the feature dimension of each head. For product $i$, let
$\sT_i\subseteq\{1,\ldots,T\}$ be its text-token index set; these sets are
mutually disjoint across products. Once $\mathbf{A}$ has been computed, value
aggregation and the output projection are linear with respect to these token
sets. The residual matrix of product $i$ is therefore
\begin{equation}
\mR_i=\left(
\operatorname{Concat}_{h=1}^{H}
\left[\sum_{j\in\sT_i}\mathbf{A}_{h,:,j}\mathbf{V}_{h,j,:}^{\top}\right]
\right)\mW_O,
\qquad
\mY=\mU+\sum_{i=1}^{N}\mR_i.
\label{eq:rsc_decomposition}
\end{equation}
Here, $\mR_i$ is the additive image-feature update induced by product $i$'s
text tokens. The matrix $\mW_O$ denotes the output projection without its
bias, while $\mU$ collects the unchanged terms, including the image-to-image
contribution, non-product text residuals, and output bias. All product
residuals are thus obtained from the same attention computation and share a
consistent scene context.

\paragraph{Target-share reweighting.}
Directly multiplying each residual $\mR_i$ by its target share $s_i$
generally fails to achieve the desired proportions, since the residuals
may differ in magnitude. We therefore define each product's current
share of the total residual magnitude as
\begin{equation}
    p_i
    = \frac{\lVert \mR_i\rVert_F}
    {\sum_{j=1}^{N}\lVert \mR_j\rVert_F}.
    \label{eq:rsc_current_share}
\end{equation}
We then compensate for this imbalance before combining the updates:
\begin{equation}
    \widetilde{\mR}_i=\frac{s_i}{p_i}\mR_i,
    \qquad
    \widetilde{\mY}
    =\mU+\sum_{i=1}^{N}\widetilde{\mR}_i.
    \label{eq:rsc_reweighting}
\end{equation}
Provided every product with a positive target share has a nonzero
residual, this rescaling makes each product's share of the summed
residual norms equal to $s_i$, while preserving the total residual
magnitude and leaving $\mU$ unchanged.

\paragraph{Share-Conditioned Prompting (SCP).}
Although RSC can modulate relative product exposure, the viewpoint and layout induced by a fixed prompt may remain largely unchanged as shares vary. This limits the effectiveness of share adjustments, leaving some products insufficiently visible even at high shares and others overly prominent even at low shares. We therefore augment each product description with prominence instructions based on its target share relative to equal allocation:
\begin{equation}
    r_i=\frac{s_i}{1/N}=Ns_i.
    \label{eq:relative-share}
\end{equation}
Here, $N$ is the number of products, and $r_i$ measures how much exposure
product $i$ is intended to receive relative to an equal share. We map $r_i$ to one of five ordered prominence levels, each associated with a shared descriptive template. Templates for higher prominence levels encourage more favorable views and greater product visibility, whereas those for lower prominence levels instruct the model to make the product less visually prominent. Because SCP maps target shares to five ordered levels, it provides discrete prompt conditioning, while RSC separately uses the real-valued target shares to calibrate residual-norm contributions.
Appendix~\ref{app:share-conditioned} provides the detailed prompt
templates. Our ablation study further demonstrates the complementary effects
of SCP and RSC in improving share--CTI monotonicity.

\subsection{Evaluating Product-Level Click-Through Intent}
\label{sec:cti}

Different levels of product exposure are expected to elicit varying degrees of user click intent, motivating the need for a click intent estimator. While conventional ad placements can draw on historical click-through rates, newly generated images require a content-based estimate of this intent. We therefore introduce a \emph{click-through intent} (CTI) proxy based on the following dimensions.

\paragraph{Evaluation dimensions.}
We focus on image-observable factors that may influence CTI while abstracting away subjective product preferences. We define four complementary dimensions:
\emph{product prominence}, which captures visual salience;
\emph{brand recognizability}, which reflects the visibility of brand cues;
\emph{scene integration}, which measures how naturally the product fits into the scene; and
\emph{contextual value communication}, which assesses whether the scene conveys the product's use, relevance, or value.
Appendix~\ref{app:cti-provement} provides further details, including an analysis of the completeness of these dimensions for CTI evaluation.

\paragraph{Image-grounded CTI evaluation.}
Given an image $I$, its scene description, and the advertised
products, a vision--language model (VLM) extracts a structured
caption documenting each product's observable evidence across
the four dimensions. Using the caption and the same scene
and product information, a separate language model (LLM)
assigns scores $z_{id}\in\{1,\ldots,5\}$ to product $i$ on
dimension $d$ under a fixed rubric; higher scores are better.
The prompt used by the VLM and LLM are provided in Appendix~\ref{app:cti}.
The equally weighted mean is linearly rescaled to $[0,1]$:
\begin{equation}
    \operatorname{CTI}_i(I)
    =\frac{1}{16}\sum_{d=1}^{4}(z_{id}-1)\in[0,1].
    \label{eq:image-cti}
\end{equation}
\subsection{Learnable Affine-Maximizer Auction Mechanism}
\label{sec:method}
We formulate exposure trading as an auction over generation configurations
subject to joint feasibility constraints.
We adopt a learnable affine maximizer auction (AMA)
\citep{curry2023differentiable} to optimize expected platform revenue, the payments from advertisers.
The mechanism selects a single common configuration using learnable
advertiser weights and configuration offsets and determines
advertiser-specific payments according to the affine-maximizer payment rule.

\paragraph{Configurations and private valuations.}
The public configuration set is indexed by
$\mathcal K=\{0,1,\ldots,K\}$. Each configuration $k\geq1$
specifies a target share vector $\mathbf s_k\in\Delta^{n-1}$ and
an expected CTI vector estimated from a historical scene set $\mathcal X$:
\begin{equation}
    q_{ik}=\frac{1}{|\mathcal X|}\sum_{x\in\mathcal X}
    \operatorname{CTI}_i(I_k(x)),
    \qquad \mathbf q_k=(q_{1k},\ldots,q_{nk})\in[0,1]^n.
    \label{eq:ama-expected-cti}
\end{equation}
Here $I_k(x)$ is a historical image generated under configuration $k$ and scene $x$.
These share and CTI vectors are fixed before bidding.
Configuration $0$ corresponds to no allocation, with
$\mathbf s_0=\mathbf q_0=\mathbf0$.

The expected CTI $q_{ik}$ captures the anticipated click intent for advertiser $i$ under configuration $k$, but need not fully characterize the commercial value of that exposure. Prior research shows that click and conversion performance need not align across ad presentations \citep{schwartz2017customer,li2025impact}. We therefore allow advertiser $i$'s private value per CTI unit to vary across configurations, reflecting its private assessments of conversion potential and campaign objectives.
Specifically, we represent
advertiser $i$'s private values per CTI unit by
a vector $\mathbf v_i=(v_{i1},\ldots,v_{iK})\in\mathbb R_+^K$, with
$v_{i0}=0$.
The resulting expected value of configuration $k$ to advertiser
$i$ is $V_i(k)=q_{ik}v_{ik}$. We assume that the platform has access to the joint distribution $F$ of advertisers' private valuations.

\paragraph{Selecting a common configuration.}
To define a learnable configuration-selection rule, we introduce trainable parameters $\phi=(\mathbf w,\boldsymbol\beta)$, where $w_i>0$ is an advertiser weight and $\beta_k$ is a configuration offset, with $\beta_0=0$. Advertiser $i$ reports its value per CTI unit
for each configuration, denoted by $b_{ik}$, so that $q_{ik}b_{ik}$
is its reported total value. We score and select configurations by
\begin{equation}
    S_\phi(k;\mathbf b)
    =\beta_k+\sum_{i=1}^n w_i q_{ik}b_{ik},
    \qquad
    k_\phi(\mathbf b)\in
    \arg\max_{k\in\mathcal K}S_\phi(k;\mathbf b).
    \label{eq:ama-score}
\end{equation}
The platform learns advertiser weights and configuration offsets
offline and fixes them before bidding. During the auction, it selects
the highest-scoring configuration, breaking ties by choosing the
first maximizer in a fixed order. If configuration $0$ is selected, no allocation is made and all
realized payments are zero, as shown in Appendix~\ref{app:ama-payments}.

\paragraph{Advertiser-specific prices and truthful incentives.}
To determine advertiser \(i\)’s payment, the platform removes its own
reported contribution from each configuration's joint score,
retaining only the offset and the other advertisers' weighted
reported values. Denote this remaining score by $C_{ik}$ and
its maximum across configurations by $H_i$:
\begin{equation}
    C_{ik}(\mathbf b_{-i})
    =\beta_k+\sum_{j\ne i}w_jq_{jk}b_{jk},
    \qquad
    H_i(\mathbf b_{-i})
    =\max_{\ell\in\mathcal K}C_{i\ell}(\mathbf b_{-i}).
    \label{eq:ama-opponent-score}
\end{equation}
The platform assigns advertiser $i$ a price for each configuration
and charges only the price of the selected configuration:
\begin{equation}
    P_{ik}(\mathbf b_{-i})
    =\frac{H_i(\mathbf b_{-i})-C_{ik}(\mathbf b_{-i})}{w_i},
    \qquad
    p_i(\mathbf b)
    =P_{i,k_\phi(\mathbf b)}(\mathbf b_{-i}).
    \label{eq:ama-payment}
\end{equation}
Because these prices depend only on the other advertisers' reports,
advertiser $i$ cannot change its own prices by changing its
report. Substituting the prices into its reported utility gives
\begin{equation}
    \begin{gathered}
        q_{ik}b_{ik}-P_{ik}(\mathbf b_{-i})
        =\frac{S_\phi(k;\mathbf b)-H_i(\mathbf b_{-i})}{w_i},
        \\[4pt]
        \arg\max_{k\in\mathcal K}
        \{q_{ik}b_{ik}-P_{ik}(\mathbf b_{-i})\}
        =\arg\max_{k\in\mathcal K}S_\phi(k;\mathbf b).
    \end{gathered}
    \label{eq:ama-menu-alignment}
\end{equation}

Consequently, the highest-scoring configuration maximizes each advertiser's reported utility, and the tie-breaking rule ensures a consistent selection. Since each advertiser's induced menu is independent of its own report, truthful reporting maximizes its true utility, establishing DSIC~\citep{hammond1979straightforward,wang2024gemnet,wang2025bundleflow}. Furthermore, every menu contains at least one zero-priced entry with nonnegative value, guaranteeing nonnegative utility under truthful reporting and hence IR.

\paragraph{Revenue-oriented learning.}
We learn weights and offsets to maximize expected revenue under
truthful reporting:
\begin{equation}
    \max_\phi R(\phi),
    \qquad
    R(\phi)=\mathbb E_{\mathbf v\sim F}
    \left[\sum_{i=1}^n p_i(\mathbf v)\right],
    \label{eq:ama-revenue-objective}
\end{equation}
where $F$ is the joint distribution of private value vectors.
To obtain gradients through discrete selection, we introduce
\begin{equation}
    \pi_\phi^\tau(k\mid\mathbf v)
    =\frac{\exp(S_\phi(k;\mathbf v)/\tau)}
    {\sum_{\ell\in\mathcal K}\exp(S_\phi(\ell;\mathbf v)/\tau)}
    \label{eq:ama-soft-selection}
\end{equation}
and use the minibatch revenue surrogate
\begin{equation}
    \mathcal L_{\mathrm{rev}}(\phi)
    =-\frac{1}{B}\sum_{m=1}^{B}\sum_{k=1}^{K}
    \pi_\phi^\tau(k\mid\mathbf v^{(m)})
    \sum_{i=1}^{n}P_{ik}(\mathbf v_{-i}^{(m)}).
    \label{eq:ama-training-loss}
\end{equation}


\section{Experiments}

\subsection{Monotonicity between Target Exposure Shares and CTI}

\paragraph{Effectiveness of our method.}
Effective exposure control should translate higher target shares into
more effective exposure in generated images, as measured by CTI.
A stronger monotonic relationship therefore indicates that the
generation method more consistently realizes the intended allocation.
Therefore, we evaluate this relationship on 1,000 two-product scenes covering
400 distinct products, generating each scene under 19 share
configurations for a total of 19,000 images. For our method, we use FLUX.1-schnell to generate images at $1024\times1024$ resolution with 10 denoising steps, applying RSC during the first 8 steps.
We compare against \citet{sun2025bidding}, which weights separately
computed generation directions by bids, and prompt-only FLUX,
which specifies exposure shares directly in the prompt.
Our method performs best across all five metrics
(Table~\ref{tab:monotonicity}), whose definitions are provided
in Appendix~\ref{app:five-metrics}.


\begin{table}[H]
\centering
\caption{Monotonicity between target exposure shares and CTI.}
\label{tab:monotonicity}
\begin{tabular}{lccccc}
\toprule
Method & Pearson $\uparrow$ & Spearman $\uparrow$ & Pairwise $\uparrow$ & Kendall's $\tau_b$ $\uparrow$ & Isotonic $R^2$ $\uparrow$ \\
\midrule
FLUX (Prompt-only)
& 0.2181 & 0.2344 & 0.5727 & 0.1837 & 0.2802 \\

Sun et al. (2025)
& 0.6675 & 0.6630 & 0.7343 & 0.5332 & 0.7708 \\

Ours
& \textbf{0.7222}
& \textbf{0.7370}
& \textbf{0.7603}
& \textbf{0.6027}
& \textbf{0.8126} \\
\bottomrule
\end{tabular}
\end{table}

\begin{figure}[H]
    \centering
    \begin{subfigure}[t]{0.48\linewidth}
        \centering
        \includegraphics[width=\linewidth]{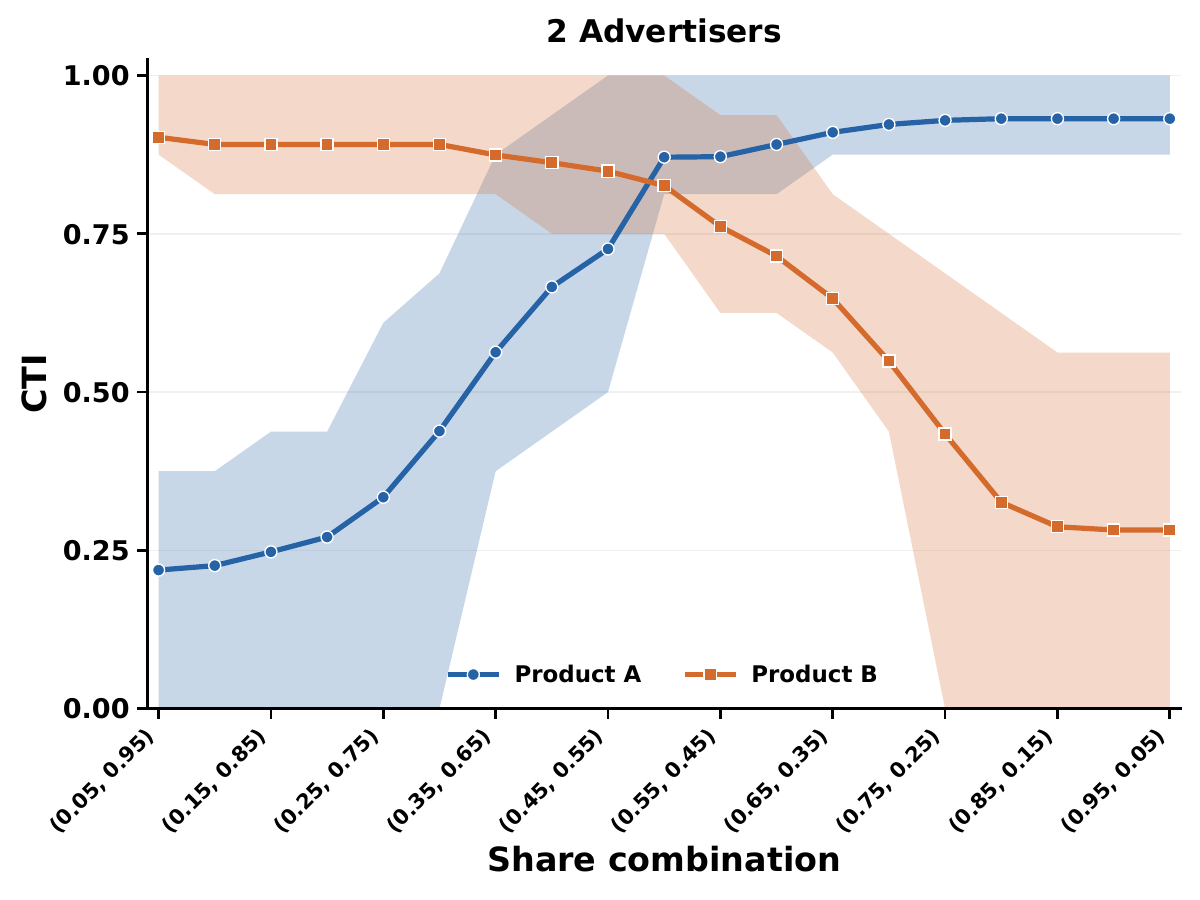}
        \caption{Two advertisers.}
        \label{fig:share-cti-two}
    \end{subfigure}
    \hfill
    \begin{subfigure}[t]{0.48\linewidth}
        \centering
        \includegraphics[width=\linewidth]{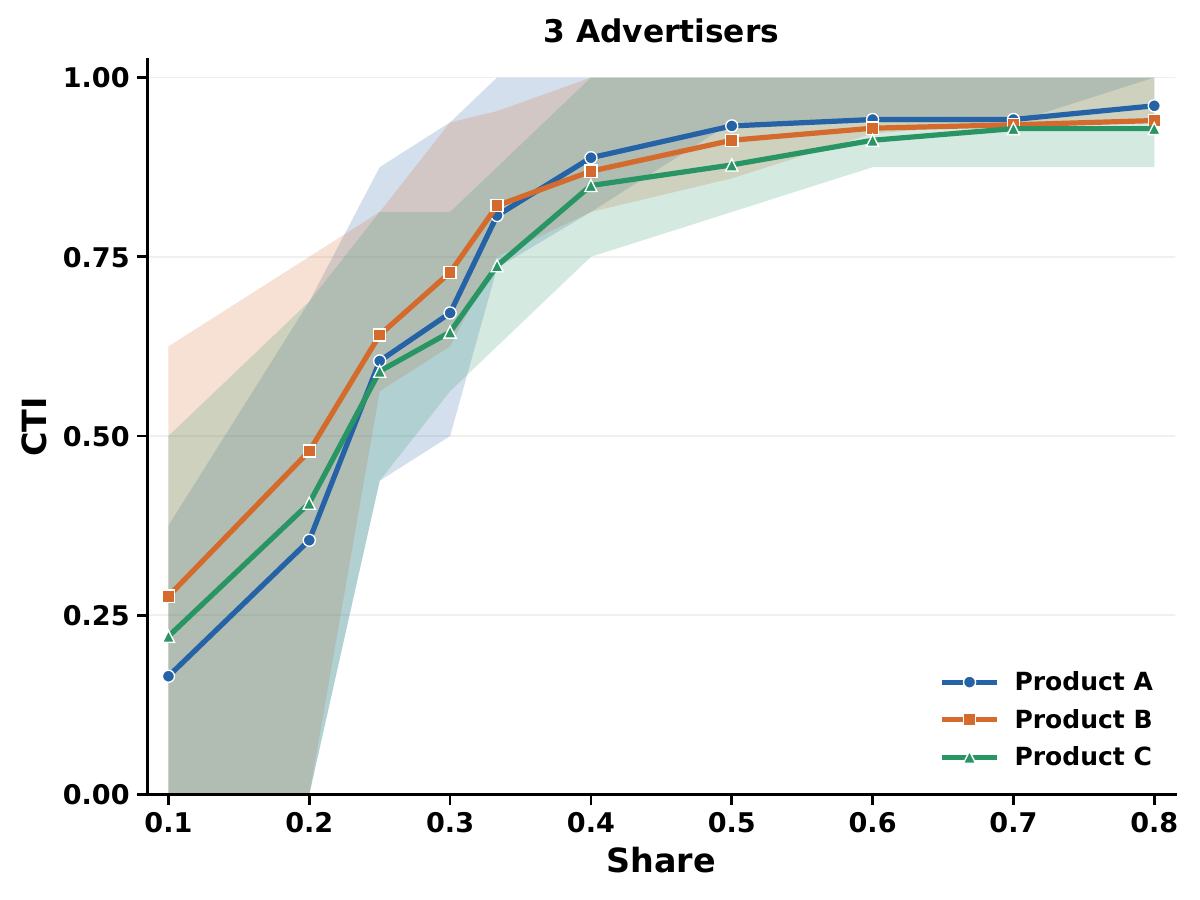}
        \caption{Three advertisers.}
        \label{fig:share-cti-three}
    \end{subfigure}
    \caption{Monotonicity between target share and CTI achieved by our method.}
    \label{fig:share-cti-curves}
\end{figure}

We further visualize the average CTI under different target-share
configurations for both the two- and three-advertiser settings in
Figure~\ref{fig:share-cti-curves}. The results are aggregated over 1,000 scenarios. The curves show a clear increasing trend as each product's target share increases. These results show that varying the target shares produces consistent changes in the effectiveness of the resulting product exposure, as measured by CTI, thereby yielding a controllable set of generation configurations for the subsequent auction.

\paragraph{Ablation study of our method.}
We evaluate the complementarity of RSC and SCP across 200 scenes,
each with 19 share configurations ranging from $(0.95, 0.05)$ to
$(0.05, 0.95)$, as shown in Table~\ref{tab:rsc-scp-ablation}.
The results show that RSC and SCP are complementary, with their
combination achieving the best performance across all metrics.
Moreover, RSC + SCP outperforms the baseline equipped with the same
SCP conditioning, indicating that the improvement cannot be attributed
to share-conditioned prompting alone.

\begin{table}[ht]
\centering
\caption{Ablation study of the monotonic relationship between target exposure shares and product-level CTI across different method configurations, evaluated on 200 scenes.}
\label{tab:rsc-scp-ablation}
\small
\setlength{\tabcolsep}{3.2pt}
\resizebox{0.9\linewidth}{!}{
\begin{tabular}{l|ccccc}
\toprule
Method
& Pearson $\uparrow$
& Spearman $\uparrow$
& Pairwise $\uparrow$
& Kendall's $\tau_b$ $\uparrow$
& Isotonic $R^2$ $\uparrow$ \\
\midrule

SCP only
& 0.3039
& 0.3172
& 0.5989
& 0.2430
& 0.4897 \\

RSC only
& 0.3730
& 0.3774
& 0.6236
& 0.2990
& 0.4964 \\

Baseline + SCP
& 0.7651
& 0.7837
& 0.7797
& 0.6430
& 0.8880 \\

RSC + SCP
& \textbf{0.8120}
& \textbf{0.8346}
& \textbf{0.8024}
& \textbf{0.6981}
& \textbf{0.9265} \\

\bottomrule
\end{tabular}
}
\end{table}

\paragraph{Efficiency Comparison.}
We compare the efficiency of our method with the baseline \citep{sun2025bidding} under generation settings involving different numbers of products. We evaluate both methods using the same set of 200 prompts and measure their inference time per denoising timestep. All inference times are measured on one NVIDIA A800 GPU and
reported as mean $\pm$ standard deviation over ten runs. As reported in Table~\ref{tab:speed_scaling}, our method is consistently faster across all product counts, with its efficiency advantage becoming increasingly pronounced as the number of products grows.

\vspace{-0.5em}
\begin{table}[H]
\centering
\caption{
Per-timestep inference-time comparison.
}
\label{tab:speed_scaling}
\setlength{\tabcolsep}{5pt}
\begin{tabular}{c cc cc}
\toprule
\# Products
& Ours (s) $\downarrow$
& \citet{sun2025bidding} (s) $\downarrow$
& Speedup $\uparrow$
& Time reduction $\uparrow$ \\
\midrule
2
& $0.6654 \pm 0.0002$
& $1.2734 \pm 0.0006$
& $1.91\times$
& 47.74\% \\
4
& $0.7317 \pm 0.0002$
& $1.6992 \pm 0.0004$
& $2.32\times$
& 56.94\% \\
6
& $0.7985 \pm 0.0018$
& $2.5485 \pm 0.0001$
& $3.19\times$
& 68.67\% \\
8
& $0.8623 \pm 0.0003$
& $3.3980 \pm 0.0017$
& $3.94\times$
& 74.62\% \\
10
& $0.9289 \pm 0.0026$
& $4.2571 \pm 0.0012$
& $4.58\times$
& 78.18\% \\
\bottomrule
\end{tabular}
\end{table}

\subsection{Validation of the CTI Evaluator}
\label{sec:cti-validation}

The CTI evaluator provides the product-level estimates used to
construct the public expected-CTI table. The auction combines this
table with advertisers' private values for allocation
and pricing. We evaluate its agreement with
human judgments and assess the reliability of the human annotations.
\paragraph{Human annotation protocol.}
We recruited 10 trained annotators to evaluate 1,000 generated images using the same four-dimensional rubric as our CTI evaluator, with each annotator rating 100 images. Product order was randomized, and target exposure shares were hidden. Annotators scored each product based on the generated image, product descriptions, and scene context. Using these human annotations as a reference, we compare
two evaluation pipelines: VLM, which directly scores
images with InternVL3-14B~\citep{zhu2025internvl3}, and VLM--LLM, which uses
InternVL3-14B to extract visual evidence and
Qwen2.5-7B-Instruct~\citep{yang2024qwen25} to score it.
Human and predicted CTI use the same aggregation and normalization.
As shown in Table~\ref{tab:cti-human-agreement},
VLM--LLM achieves higher agreement with human judgments
across all three metrics,
supporting its use as our CTI evaluator.

\begin{table}[H]
    \centering
    \setlength{\tabcolsep}{3.5pt}
    \caption{Agreement between predicted CTI and human judgments
    on 1,000 images from 200 scenes. Results are averaged over
    the two products. Higher values are better.}
    \label{tab:cti-human-agreement}
    \begin{tabular}{lccc}
        \hline
        Evaluator
        & Pearson $\uparrow$
        & Spearman $\uparrow$
        & ICC(2,1) $\uparrow$ \\
        \hline
        VLM only
        & 0.662 & 0.574 & 0.610 \\
        VLM + LLM (ours)
        & \textbf{0.669} & \textbf{0.603} & \textbf{0.653} \\
        \hline
    \end{tabular}
\end{table}

\paragraph{Inter-annotator agreement.}
Table~\ref{tab:human_agreement} reports agreement among five annotators on 20 scenes to examine the reliability of human ratings.
Ordinal Krippendorff's $\alpha$ measures chance-corrected
agreement while accounting for rating order.
Exact agreement is the percentage of evaluated items for which all five annotators assign exactly the same score on a given metric. MAE measures the average absolute difference
between annotator pairs. All metrics achieve ordinal $\alpha>0.8$, indicating high inter-annotator agreement. Although exact agreement on CTI is only 42.0\% under the strict requirement that all five annotators assign identical scores, its low MAE of 0.0450 indicates that disagreements are small on average. These results support the reliability of human ratings.

\begin{table}[H]
\centering
\caption{Inter-annotator agreement among 20 scenes. Each image was independently rated by five annotators. We calculated agreement among the annotators’ ratings using all 100 images from the 20 scenes.
PP, BR, SI, and CVC denote four dimensions.}
\label{tab:human_agreement}
\begin{tabular}{lccccc}
\toprule
Measure & PP & BR & SI & CVC & CTI \\
\midrule
Ordinal $\alpha$ $\uparrow$ & 0.8181 & 0.9654 & 0.8342 & 0.8419 & 0.9179 \\
Exact (\%) $\uparrow$      & 78.0   & 82.0   & 66.0   & 68.0   & 42.0   \\
MAE   $\downarrow$           & 0.1600 & 0.1400 & 0.2933 & 0.2733 & 0.0450 \\
\bottomrule
\end{tabular}
\end{table}

\subsection{Evaluating the Economic Performance of the Auction Mechanism}
\label{sec:exp-tcom}

We evaluate whether learning advertiser weights and configuration offsets
improves platform revenue under different structures of private preferences.
We conduct separate experiments with two and three advertisers, each using
a fixed set of 19 paid configurations and a no-allocation option.
Expected CTI is averaged across historical scenes and fixed across
valuation profiles, with total configuration value given by
$q_{ik}v_{ik}$.

\paragraph{Valuation distributions and baselines.}
We consider two synthetic valuation distributions. In the
\emph{correlated} setting, we independently sample two endpoint
coefficients from $\mathrm{Uniform}[0,1]$ for each advertiser and
interpolate its values per CTI unit linearly over exposure shares.
This models preferences that vary smoothly across nearby exposure
configurations. In the \emph{independent} setting, we sample every
$v_{ik}$ independently from $\mathrm{Uniform}[0,1]$, allowing
configuration-specific preferences without the interpolation constraint.
We train AMA separately for each setting and select among three
training seeds using validation revenue. Revenue is the mean sum of actual payments under truthful reporting
and hard configuration selection. We compare against VCG~\citep{vickrey1961counterspeculation,
clarke1971multipart,groves1973incentives} and a Myerson-based baseline~\citep{myerson1981optimal}. VCG maximizes
total reported value and uses pivot payments. The Myerson-based
baseline uses unit advertiser weights and fixed offsets
$\beta_k=-\tfrac12\sum_i q_{ik}$ with AMA payments, retaining the
uniform single-parameter Myerson scoring rule without claiming
optimality in the present multidimensional setting.

\begin{table}[H]
    \centering
    \setlength{\tabcolsep}{6pt}
    \caption{Mean platform revenue under truthful reporting.}
    \label{tab:ama-revenue}
    \begin{tabular}{clccc}
        \toprule
        Advertisers & Distribution & VCG & Myerson-based & AMA \\
        \midrule
        2   & Correlated  & 0.1592 & 0.4036 & \textbf{0.4242} \\
        2   & Independent & 0.3699 & 0.5648 & \textbf{0.8535} \\
        \midrule
        3 & Correlated  & 0.2019 & 0.5238 & \textbf{0.5462} \\
        3 & Independent & 0.3498 & 0.6031 & \textbf{0.9086} \\
        \bottomrule
    \end{tabular}%
\end{table}

\paragraph{Results.}
Table~\ref{tab:ama-revenue} shows that learned AMA achieves the highest
revenue in all four settings. Under correlated valuations, it improves
over the Myerson-based baseline by 5.10\% and 4.27\% for two and three
advertisers, respectively; under independent valuations, the gains
reach 51.12\% and 50.66\%. It also consistently outperforms VCG in
platform revenue. These results demonstrate the benefit of adapting
auction parameters to configuration-dependent private valuations,
with larger gains over the fixed baselines in the independent setting.

\section{Conclusion}
\label{sec:conclusion}

We presented DiffAuction, a unified framework integrating controllable image generation, CTI evaluation, and truthful auction design. RSC and SCP enable continuous exposure control with one model forward pass per denoising step, while the CTI evaluator provides interpretable product-level click-intent estimates. A learnable affine maximizer combines expected CTI with advertisers' private values to select feasible configurations and determine payments while ensuring DSIC and IR. Experiments demonstrate improved share--CTI monotonicity, generation efficiency, human agreement, and auction revenue.


\newpage



\bibliographystyle{manuscript}
\bibliography{references}

\appendix
\newpage
\section{Further experiments on the monotonicity between target exposure share and CTI}

We further evaluate monotonicity in the three-advertiser setting and compare our method with the baseline, as shown in Table~\ref{tab:three-agent-monotonicity}. The results show that our method continues to outperform the baseline while maintaining strong monotonicity.

\begin{table}[H]
\centering
\caption{Monotonicity between target exposure shares and CTI in the three-advertiser setting.}
\label{tab:three-agent-monotonicity}
\begin{tabular}{lccccc}
\toprule
Method
& Pearson $\uparrow$
& Spearman $\uparrow$
& Pairwise $\uparrow$
& Kendall's $\tau_b$ $\uparrow$
& Isotonic $R^2$ $\uparrow$ \\
\midrule
Sun et al. (2025)
& 0.5861
& 0.6579
& 0.7530
& 0.5449
& 0.6143 \\

Ours
& \textbf{0.6839}
& \textbf{0.7932}
& \textbf{0.8171}
& \textbf{0.6685}
& \textbf{0.6922} \\
\bottomrule
\end{tabular}
\end{table}

\begin{figure}[H]
    \centering
    \includegraphics[width=1.0\linewidth]{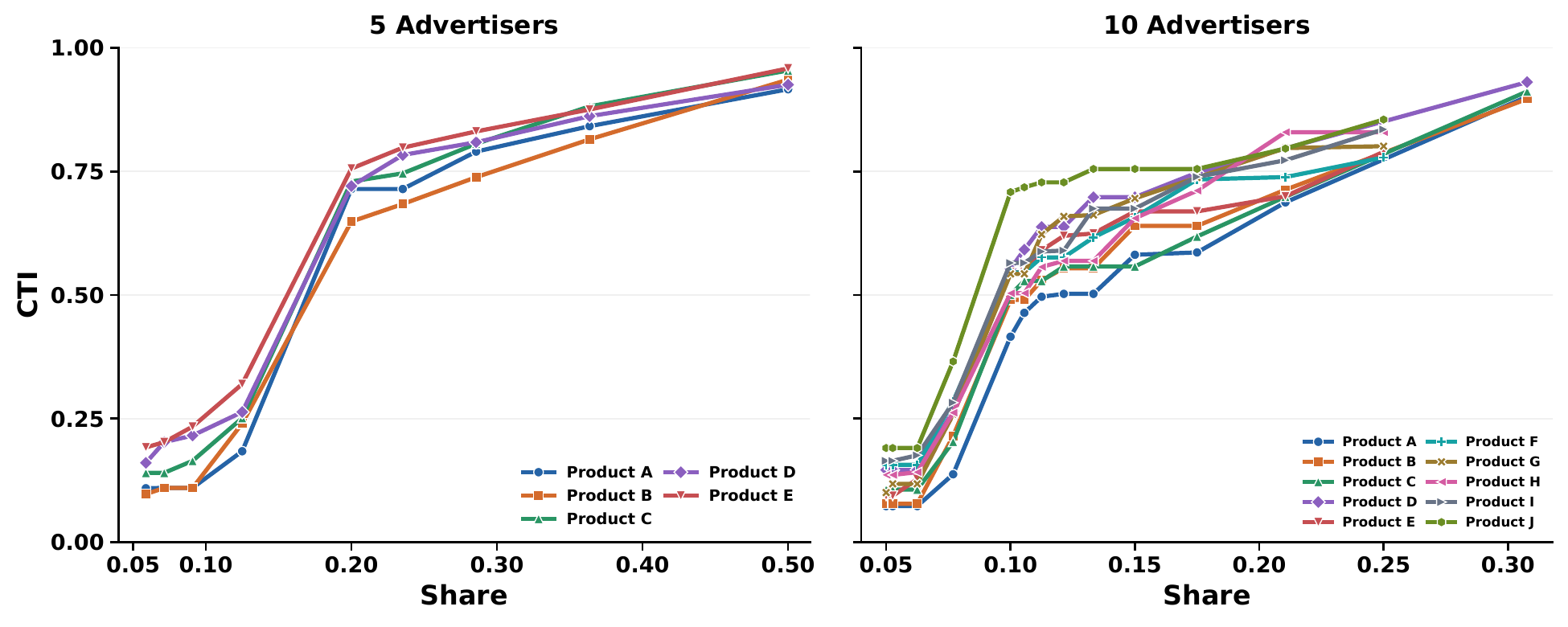}
    \caption{Monotonicity between target exposure share and CTI under extreme numbers of competing products.}
    \label{fig:5-10}
\end{figure}

We further increase the number of advertisers to five and ten to examine whether the monotonic relationship can still be maintained. We use 100 scene prompts and generate 2,000 images for evaluation. The results are reported in Table~\ref{tab:multi-product-monotonicity} and Figure~\ref{fig:5-10}. The results show that even when a larger number of products compete for exposure simultaneously, our method still preserves a clear monotonic relationship between target shares and the resulting CTI, demonstrating its scalability to settings with more advertisers.

\begin{table}[H]
\centering
\caption{Monotonicity under increasing numbers of advertised products.}
\label{tab:multi-product-monotonicity}
\begin{tabular}{cccccc}
\toprule
Setting
& Pearson $\uparrow$
& Spearman $\uparrow$
& Pairwise $\uparrow$
& Kendall's $\tau_b$ $\uparrow$
& Isotonic $R^2$ $\uparrow$ \\
\midrule
5 products
& 0.7084
& 0.7324
& 0.7773
& 0.6037
& 0.6992 \\

10 products
& 0.5928
& 0.6004
& 0.7162
& 0.4981
& 0.5689 \\
\bottomrule
\end{tabular}
\end{table}

In addition, we investigate whether our method preserves monotonicity when the competing products belong to the same category, such as a Luckin coffee cup and a Starbucks coffee cup. We construct 50 scenarios involving competition between products from the same category and generate 1,000 images in total, with two advertisers competing in each image. We then evaluate the monotonicity between CTI and exposure share. The results are shown in Table~\ref{tab:competing-product-monotonicity}.



\begin{table}[H]
    \centering
    \small
    \setlength{\tabcolsep}{6pt}
    \caption{Share--CTI monotonicity for competing products.}
    \label{tab:competing-product-monotonicity}
    \begin{tabular}{lccccc}
        \toprule
        Method
        & Pearson $\uparrow$
        & Spearman $\uparrow$
        & Pairwise $\uparrow$
        & Kendall's $\tau_b$ $\uparrow$
        & Isotonic $R^2$ $\uparrow$ \\
         \midrule
        \citet{sun2025bidding}
        & 0.4794
        & 0.4858
        & 0.6597
        & 0.4007
        & 0.6687 \\
        Ours
        & \textbf{0.5600}
        & \textbf{0.5961}
        & \textbf{0.7073}
        & \textbf{0.4964}
        & \textbf{0.6755} \\
        \bottomrule
    \end{tabular}
\end{table}

\section{Additional Details on Share Control}

\label{app:share-control}

\subsection{Further Experiments on RSC}


\paragraph{Number of controlled denoising steps.}
We vary the number of denoising steps during which RSC is applied to study the
trade-off between control effectiveness and computational cost. For each
setting, we report the monotonicity between the target exposure share and CTI.
As shown in Table~\ref{tab:rsc-control-steps}, increasing the control duration
from two to eight steps consistently improves all five metrics, indicating
that sustained residual calibration strengthens share--CTI monotonicity
over the evaluated configurations. The best performance is
obtained with eight controlled steps. Extending RSC to all ten denoising steps
leads to a slight degradation, suggesting that intervention during the final
refinement stage may over-constrain local appearance and disrupt the
previously established prominence ordering. Therefore, controlling the first
eight steps provides the best balance between monotonicity and computational
overhead.

\begin{table}[H]
    \centering
    \small
    \caption{Effect of the number of controlled denoising steps on
    share--CTI monotonicity and inference time. Runtime is averaged over three runs on an NVIDIA A800 GPU.}
    \label{tab:rsc-control-steps}
    \setlength{\tabcolsep}{3.2pt}
    \begin{tabular}{ccccccc}
        \toprule
        Control Steps
        & Pearson $\uparrow$
        & Spearman $\uparrow$
        & Pairwise $\uparrow$
        & Kendall's $\tau_b$ $\uparrow$
        & Isotonic $R^2$ $\uparrow$
        & Runtime (s) $\downarrow$ \\
        \midrule
        2
        & 0.7581 & 0.7839 & 0.7788 & 0.6413 & 0.8604
        & $\mathbf{5.448 \pm 0.003}$ \\
        4
        & 0.7875 & 0.7925 & 0.7823 & 0.6522 & 0.8972
        & $6.422 \pm 0.002$ \\
        6
        & 0.7892 & 0.8074 & 0.7920 & 0.6717 & 0.8979
        & $7.401 \pm 0.002$ \\
        8
        & \textbf{0.8120}
        & \textbf{0.8346}
        & \textbf{0.8024}
        & \textbf{0.6981}
        & \textbf{0.9265}
        & $8.378 \pm 0.006$ \\
        10
        & 0.8083 & 0.8249 & 0.7987 & 0.6893 & 0.9184
        & $9.344 \pm 0.001$ \\
        \bottomrule
    \end{tabular}
\end{table}

\paragraph{Intervention location.}
We further investigate where RSC should be applied within FLUX. Specifically, we compare three variants: RSC applied only to the double-stream blocks, only to the single-stream blocks, and to both types of blocks. This analysis examines how the intervention stage affects the resulting share--CTI monotonicity. All variants are evaluated on the same 200 scenes. As shown in
Table~\ref{tab:block-ablation}, applying RSC to both the double-stream and
single-stream blocks achieves the best performance across all metrics. This is
because the double-stream blocks establish early interactions between product
tokens and image features, whereas the subsequent single-stream blocks further
integrate and refine these multimodal representations. Applying RSC throughout
both stages therefore preserves the intended product-wise residual allocation
during early alignment and reinforces it during later feature refinement,
leading to more consistent share control.

\begin{table}[H]
    \centering
    \small
    \setlength{\tabcolsep}{5pt}
    \caption{Effect of applying residual control to different FLUX transformer streams.}
    \label{tab:block-ablation}
    \begin{tabular}{lccccc}
        \toprule
        Controlled Blocks
        & Pearson $\uparrow$
        & Spearman $\uparrow$
        & Pairwise $\uparrow$
        & Kendall's $\tau_b$ $\uparrow$
        & Isotonic $R^2$ $\uparrow$ \\
        \midrule
        Double-stream only
        & 0.6878 & 0.7021 & 0.7444 & 0.5649 & 0.7861 \\
        Single-stream only
        & 0.6478 & 0.6605 & 0.7297 & 0.5291 & 0.7588 \\
        Double- and single-stream
        & \textbf{0.8120} & \textbf{0.8346} & \textbf{0.8024}
        & \textbf{0.6981} & \textbf{0.9265} \\
        \bottomrule
    \end{tabular}
\end{table}

\paragraph{Ablation of residual reweighting.}
We examine whether normalizing product-specific residual
magnitudes improves share control compared with directly
scaling residuals by their target shares. Direct scaling
uses $\widetilde{R}_i=s_iR_i$, whereas RSC applies
$\widetilde{R}_i=(s_i/p_i)R_i$, where
$p_i=\|R_i\|_F/\sum_j\|R_j\|_F$.
The latter redistributes residual magnitudes according to
the target shares while preserving the sum of their
Frobenius norms. Table~\ref{tab:residual-reweighting-ablation}
compares the two strategies in terms of share--CTI
monotonicity. The results show that RSC achieves better performance. This is primarily because direct scaling does not ensure that each product’s share of the summed residual norms matches its target share. It also reduces the sum of the product-specific residual norms, which may weaken product visibility. RSC compensates for differences in residual magnitudes while preserving the sum of their Frobenius norms.

\begin{table}[H]
    \centering
    \caption{Comparison of direct residual scaling and RSC.}
    \label{tab:residual-reweighting-ablation}
    \small
    \begin{tabular}{lccccc}
        \toprule
        Method
        & Pearson $\uparrow$
        & Spearman $\uparrow$
        & Pairwise $\uparrow$
        & Kendall's $\tau_b$ $\uparrow$
        & Isotonic $R^2$ $\uparrow$ \\
        \midrule
        Direct scaling
        & 0.7792 & 0.7705 & 0.7687 & 0.6232 & 0.8393 \\
        RSC
        & \textbf{0.8120} & \textbf{0.8346} & \textbf{0.8024}
        & \textbf{0.6981} & \textbf{0.9265} \\
        \bottomrule
    \end{tabular}
\end{table}

\subsection{Definitions of Correlation and Monotonicity Metrics}
\label{app:five-metrics}
In this section, we introduce five metrics for evaluating the monotonicity between exposure share and CTI. We describe their meanings and calculation methods below.

\paragraph{Pearson correlation.}
Pearson correlation measures the strength of the
\emph{linear relationship} between target share and CTI:
\begin{equation}
r =
\frac{\sum_k(s_k-\bar{s})(y_k-\bar{y})}
{\sqrt{\sum_k(s_k-\bar{s})^2
       \sum_k(y_k-\bar{y})^2}},
\end{equation}
where $\bar{s}$ and $\bar{y}$ are the means within the current
scene--product curve. A larger value indicates a stronger
positive linear association.

\paragraph{Spearman correlation.}
Spearman correlation measures \emph{rank consistency},
allowing the share--CTI relationship to be nonlinear:
\begin{equation}
\rho =
\operatorname{Corr}
\bigl(\operatorname{rank}(\mathbf{s}),
      \operatorname{rank}(\mathbf{y})\bigr).
\end{equation}
Tied CTI values receive their average rank.
A larger value indicates that configurations with higher
target shares generally receive higher CTI.

\paragraph{Pairwise monotonicity.}
This metric measures how frequently a higher target share
corresponds to higher CTI. With $M=\binom{K}{2}$, we compute
\begin{equation}
P =
\frac{1}{M}\sum_{k<\ell}
\left[
\mathbb{I}(y_\ell>y_k)
+\frac{1}{2}\mathbb{I}(y_\ell=y_k)
\right].
\end{equation}
An increase receives one point, a tie receives half a point,
and a decrease receives zero.
Thus, $P=1$ indicates strict increase across all pairs,
whereas a constant CTI curve receives $P=0.5$.

\paragraph{Kendall's $\tau_b$.}
Kendall's $\tau_b$ measures the monotonic association between target
share and CTI while accounting for ties. For each pair of configurations
with $s_\ell > s_k$, we define the pair as concordant if
$y_\ell > y_k$, discordant if $y_\ell < y_k$, and tied if
$y_\ell = y_k$. Let $C$, $D$, and $T$ denote the numbers of
concordant, discordant, and CTI-tied pairs, respectively. Since the
target shares contain no ties, Kendall's $\tau_b$ reduces to
\begin{equation}
\tau_b =
\frac{C-D}
{\sqrt{(C+D)(C+D+T)}}.
\end{equation}
A larger $\tau_b$ indicates a stronger monotonically increasing relationship.

\paragraph{Isotonic $R^2$.}
Isotonic $R^2$ measures how closely the CTI curve can be
approximated by a nondecreasing sequence:
\begin{equation}
\widehat{\mathbf y}
=
\arg\min_{z_1\leq\cdots\leq z_K}
\sum_{k=1}^{K}(y_k-z_k)^2,
\qquad
R_{\mathrm{iso}}^2
=
1-
\frac{\sum_k(y_k-\widehat y_k)^2}
     {\sum_k(y_k-\bar y)^2}.
\end{equation}
We obtain the fit using the pool-adjacent-violators algorithm.
This metric accounts for the \emph{magnitude of departures}
from a nondecreasing curve relative to the curve's total
variation. A nonconstant, nondecreasing curve achieves one,
including curves with plateaus; therefore, a high value
does not by itself imply fine-grained share responsiveness.
\subsection{Share-conditioned prompts}
\label{app:share-conditioned}

We augment product descriptions with share-conditioned prominence instructions to address limited exposure control under a fixed viewpoint and layout. For each product $i$, we compute its target share relative to equal allocation:
\begin{equation}
    r_i = \frac{s_i}{1/N} = Ns_i,
\end{equation}
where $N$ is the number of products and $\sum_i s_i=1$. Thus, $r_i=1$ corresponds to equal allocation, providing a common reference across different numbers of products.

We map each positive relative share to one of five prominence levels:
\begin{equation}
L(r_i)=
\begin{cases}
1, & 0 < r_i < 1-\alpha,\\
2, & 1-\alpha \leq r_i < 1-\delta,\\
3, & 1-\delta \leq r_i < 1+\delta,\\
4, & 1+\delta \leq r_i \leq 1+\alpha,\\
5, & r_i > 1+\alpha,
\end{cases}
\label{eq:prominence-levels}
\end{equation}
where $0 < \delta < \alpha < 1$. We use $\alpha=0.5$ and $\delta=0.2$, yielding boundaries of $0.5$, $0.8$, $1.2$, and $1.5$.

Each level uses the following template, with \texttt{[product]} replaced by the corresponding product description:
\begin{itemize}
    \item \textbf{Level 1:} Keep [product] visible but substantially less prominent, with a smaller visual footprint and a more peripheral or background position.
    \item \textbf{Level 2:} Keep [product] less prominent but visible.
    \item \textbf{Level 3:} Keep [product] visible.
    \item \textbf{Level 4:} Frame [product] somewhat more prominently than other products.
    \item \textbf{Level 5:} The camera should face [product] directly and present it as the primary focal product.
\end{itemize}

These share-conditioned instructions encourage more favorable views of higher-share products while limiting excessive exposure of lower-share products. They complement residual control by enabling viewpoint and layout changes that residual reweighting alone may not achieve under a fixed prompt.

\paragraph{Analysis of threshold values.}
\label{app:threshold-analysis}

We further investigate how $\alpha$ and $\delta$ affect the resulting monotonicity. Intuitively, $\alpha$ controls the boundary between mild and strong prominence preferences, while $\delta$ determines the neutral region around equal allocation. We select three parameter configurations and evaluate their effects on the monotonicity metrics using the same 50 scenes.

\begin{table}[H]
    \centering
    \small
    \setlength{\tabcolsep}{3.5pt}
    \caption{Hyperparameter search for threshold values.}
    \label{tab:threshold-hyperparameter-search}
    \begin{tabular}{cccccccc}
        \toprule
        Advertisers
        & $\alpha$
        & $\delta$
        & Pearson $\uparrow$
        & Spearman $\uparrow$
        & Pairwise $\uparrow$
        & Kendall's $\tau_b$ $\uparrow$
        & Isotonic $R^2$ $\uparrow$ \\
        \midrule

        2
        & 0.40 & 0.15
        & 0.7716
        & 0.7750
        & 0.7802
        & 0.6431
        & 0.8788 \\

        2
        & 0.50 & 0.20
        & \textbf{0.7747}
        & \textbf{0.7829}
        & \textbf{0.7826}
        & \textbf{0.6462}
        & 0.8840 \\

        2
        & 0.60 & 0.25
        & 0.7626
        & 0.7706
        & 0.7758
        & 0.6333
        & \textbf{0.8886} \\

        \midrule

        3
        & 0.40 & 0.15
        & 0.6899
        & 0.7590
        & 0.7956
        & 0.6370
        & 0.6922 \\

        3
        & 0.50 & 0.20
        & \textbf{0.7042}
        & \textbf{0.7817}
        & \textbf{0.8071}
        & \textbf{0.6579}
        & \textbf{0.7048} \\

        3
        & 0.60 & 0.25
        & 0.6783
        & 0.7560
        & 0.7920
        & 0.6340
        & 0.6760 \\

        \bottomrule
    \end{tabular}
\end{table}
\subsection{Visualization of exposure-share control}
\begin{figure}[t]
    \centering
    \includegraphics[width=1.0\linewidth]{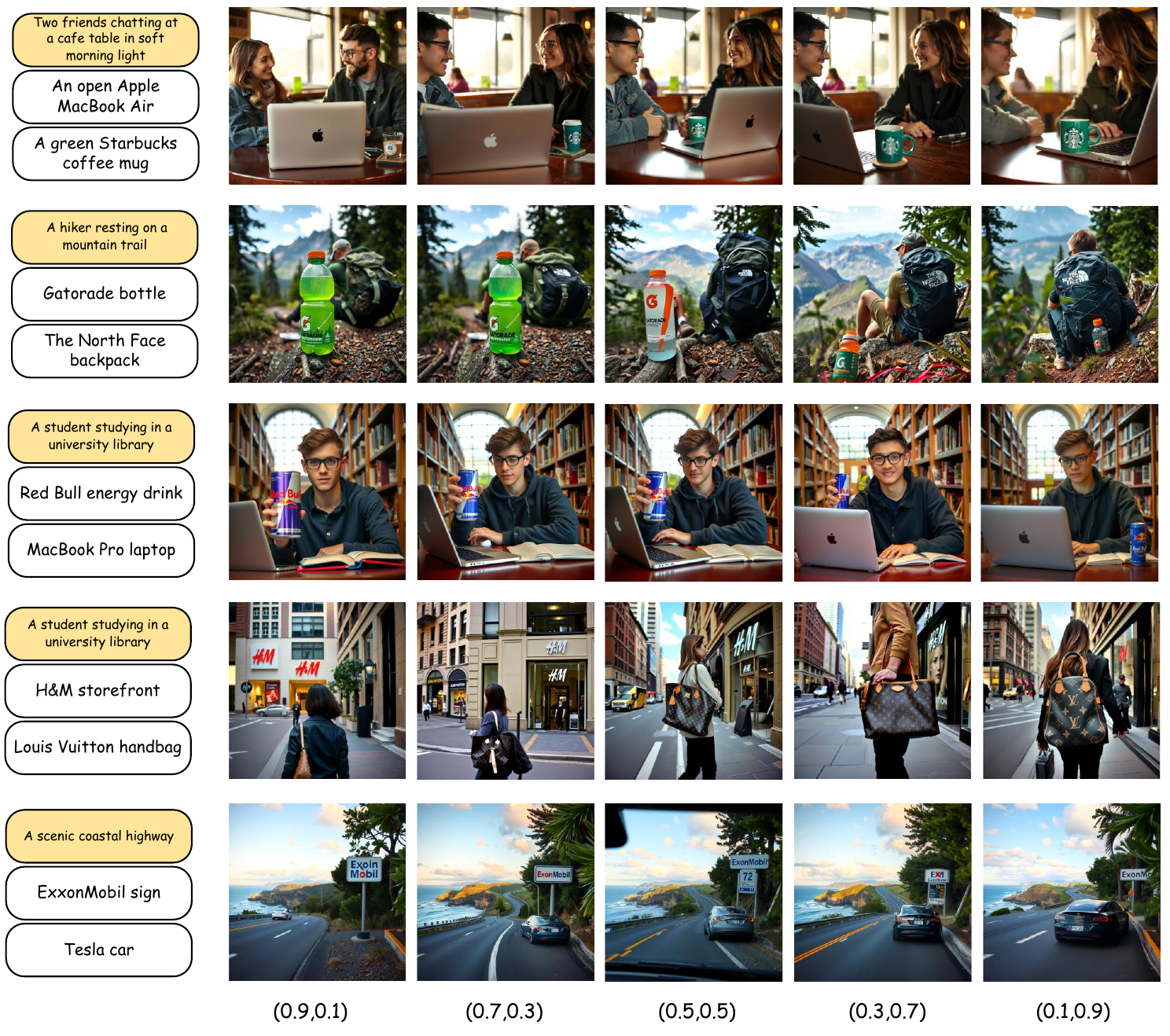}
    \caption{Visualization of exposure-share control.}
    \label{fig:vis}
\end{figure}

To illustrate the effectiveness of our method, we render five scenes, each under five evenly spaced exposure-share configurations, as shown in Figure~\ref{fig:vis}. As the target share of a product increases, its exposure in the generated image also increases accordingly.

\section{Further Experiments on CTI evaluation}


Section~\ref{sec:cti-validation} compares human and model
scores under the same four-dimensional rubric. To further
examine the relationship between CTI and human click intent,
we recruited five additional annotators who had not
participated in the earlier rubric-based evaluation.
Without access to the four-dimensional rubric, they rated
their willingness to click on the products depicted
in the images.

The evaluation set comprises 570 images from 30 scenes,
with 19 generated images per scene. The images were presented
in shuffled order, and each annotator independently rated
the complete set. For each evaluated item, we averaged
the five annotators' ratings and compared the resulting
mean click-intent score with model-predicted CTI.

Our results show that model-predicted CTI achieves Pearson
and Spearman correlations of $0.8526$ and $0.8339$,
respectively, with the mean human click-intent ratings,
indicating strong linear and rank associations.
The ICC(2,1) between model-predicted CTI and the mean human
ratings is $0.7694$. These results provide further support
for CTI as a proxy for average self-reported click intent
within the evaluated sample.

\section{Additional Details on the Auction Mechanism}
\label{app:training-details}
We train a separate AMA for each advertiser count and valuation
distribution, initializing advertiser weights to $w_i=1$ and
paid-configuration offsets to $\beta_k=-\frac12\sum_i q_{ik}$
as in the Myerson-based baseline, and subsequently learning both.
We parameterize weights as $w_i=\exp(a_i)$ and constrain $a_i$
to $[-1,1]$ and paid-configuration offsets to $[-2,1]$ after
each update, while fixing the no-allocation offset at zero.
We use Adam with a learning rate of $0.003$, a batch size of
$512$, and gradient norm clipping at $10$ for $5{,}000$ updates.
The softmax temperature decreases exponentially from $0.04$
to $0.0005$. The no-allocation outcome is included when
computing selection probabilities but contributes zero revenue.

During training, we periodically evaluate the model on a fixed
validation set of $16{,}384$ valuation profiles. For each profile,
we select the highest-scoring configuration and compute platform
revenue using the AMA payment rule, retaining the model with
the highest mean validation revenue. We evaluate the selected
model on $131{,}072$ independent test profiles, shared by all
mechanisms within each setting. All reported revenues are
computed from payments for the selected configurations,
rather than the softmax approximation used during training.

\section{Proofs of DSIC and IR}
\label{app:dsic-proof}

We establish dominant-strategy incentive compatibility (DSIC)
and individual rationality (IR) for the affine maximizer auction
defined in \eqref{eq:ama-score}--\eqref{eq:ama-payment}.
The guarantees apply to any number of advertisers and any finite
configuration set.

\paragraph{Setting and assumptions.}
Fix the public configuration set
$\mathcal K=\{0,1,\ldots,K\}$, the public expected-CTI table $q_{ik}\in[0,1]$, and the mechanism parameters
$\phi=(\mathbf w,\boldsymbol\beta)$, where
$w_i>0$, $\beta_0=0$, and $q_{i0}=0$ for every advertiser.
The configuration set, CTI table, and mechanism parameters
are fixed independently of the current reports.
All score ties are resolved using the same predetermined order
over configurations.

Advertiser $i$ has a multidimensional private type
$\mathbf v_i=(v_{i1},\ldots,v_{iK})\in\mathbb R_+^K$,
where $v_{ik}$ denotes its true value per CTI unit under
configuration $k$.
Its total expected value for that configuration is modeled as
$V_i(k)=q_{ik}v_{ik}$.
The advertiser submits a report
$\mathbf b_i=(b_{i1},\ldots,b_{iK})\in\mathbb R_+^K$.
For convenience, set $v_{i0}=b_{i0}=0$.
Utilities are quasi-linear: selecting configuration $k$
and charging payment $p_i$ gives utility
$q_{ik}v_{ik}-p_i$.

The components of $\mathbf v_i$ need not be equal,
independent, or monotone across configurations.
The proof also applies when admissible type and report vectors
belong to a restricted subset of $\mathbb R_+^K$, provided
truthful reporting is admissible.
No distributional assumption on private types is required.

\subsection{Report-Independent Menus and Common Optimality}
\label{app:ama-menu-optimality}

\paragraph{Lemma 1 (Common menu optimality).}
For every report profile $\mathbf b$ and advertiser $i$,
the price menu
$\{P_{ik}(\mathbf b_{-i})\}_{k\in\mathcal K}$
is independent of $\mathbf b_i$, and
\begin{equation}
    \arg\max_{k\in\mathcal K}
    \left\{q_{ik}b_{ik}-P_{ik}(\mathbf b_{-i})\right\}
    =
    \arg\max_{k\in\mathcal K}S_\phi(k;\mathbf b).
    \label{eq:app-ama-common-optimum}
\end{equation}
Consequently, applying the shared tie-breaking rule to
each advertiser's reported utility-maximizing entries
selects the same configuration.

\paragraph{Proof.}
Fix advertiser $i$ and the other reports $\mathbf b_{-i}$.
Recall that
\[
    C_{ik}(\mathbf b_{-i})
    =\beta_k+\sum_{j\ne i}w_jq_{jk}b_{jk},
    \qquad
    H_i(\mathbf b_{-i})
    =\max_{\ell\in\mathcal K}C_{i\ell}(\mathbf b_{-i}).
\]
Both quantities are independent of $\mathbf b_i$.
Since $w_i$ is fixed before bidding, the price
\[
    P_{ik}(\mathbf b_{-i})
    =\frac{H_i(\mathbf b_{-i})-C_{ik}(\mathbf b_{-i})}{w_i}
\]
is also independent of advertiser $i$'s own report.

For every configuration $k$,
\begin{align}
    q_{ik}b_{ik}-P_{ik}(\mathbf b_{-i})
    &=
    \frac{w_iq_{ik}b_{ik}
          +C_{ik}(\mathbf b_{-i})
          -H_i(\mathbf b_{-i})}{w_i}
    =
    \frac{S_\phi(k;\mathbf b)
          -H_i(\mathbf b_{-i})}{w_i}.
    \label{eq:app-ama-utility-alignment}
\end{align}
The term $H_i(\mathbf b_{-i})$ is constant across
configurations, and $w_i>0$.
Subtracting this constant and dividing by $w_i$
therefore preserve the ordering of configurations,
including ties.
This proves \eqref{eq:app-ama-common-optimum}.
The common tie-breaking rule selects the same element
from these identical maximizer sets.

\subsection{Dominant-Strategy Incentive Compatibility}
\label{app:dsic}

\paragraph{Proposition 1 (DSIC).}
Under the allocation and payment rules in
Equations~\eqref{eq:ama-score} and~\eqref{eq:ama-payment},
truthful reporting is a weakly dominant strategy.
For every advertiser $i$, true type $\mathbf v_i$,
alternative report $\mathbf r_i$, and profile of other
reports $\mathbf b_{-i}$, define
\[
    k^{\mathrm T}=k_\phi(\mathbf v_i,\mathbf b_{-i}),
    \qquad
    k^{\mathrm R}=k_\phi(\mathbf r_i,\mathbf b_{-i}).
\]
Then
\begin{equation}
    \begin{aligned}
    &q_{i,k^{\mathrm T}}v_{i,k^{\mathrm T}}
      -p_i(\mathbf v_i,\mathbf b_{-i})
    \\
    &\qquad\geq
      q_{i,k^{\mathrm R}}v_{i,k^{\mathrm R}}
      -p_i(\mathbf r_i,\mathbf b_{-i}).
    \end{aligned}
    \label{eq:app-dsic}
\end{equation}

\paragraph{Proof.}
Fix $i$, $\mathbf v_i$, and $\mathbf b_{-i}$.
By Lemma~1, reporting $\mathbf b_i=\mathbf v_i$
selects a configuration that maximizes advertiser $i$'s
true utility over its entire menu:
\[
    k^{\mathrm T}\in
    \arg\max_{k\in\mathcal K}
    \left\{q_{ik}v_{ik}-P_{ik}(\mathbf b_{-i})\right\}.
\]
A deviation to $\mathbf r_i$ may change the selected
configuration to $k^{\mathrm R}$, but does not change
any menu price.
Since $k^{\mathrm R}\in\mathcal K$, optimality of
$k^{\mathrm T}$ implies
\[
    q_{i,k^{\mathrm T}}v_{i,k^{\mathrm T}}
    -P_{i,k^{\mathrm T}}(\mathbf b_{-i})
    \geq
    q_{i,k^{\mathrm R}}v_{i,k^{\mathrm R}}
    -P_{i,k^{\mathrm R}}(\mathbf b_{-i}).
\]
The mechanism charges the price of the selected entry,
so this is precisely \eqref{eq:app-dsic}.

The deviation $\mathbf r_i$ may change arbitrarily
many coordinates simultaneously.
The argument does not assume that the other advertisers
report truthfully.
Multiple maximizing configurations do not affect DSIC:
they yield the same true utility, so changing between
them cannot produce a strict gain.

\subsection{Individual Rationality}
\label{app:ir}

\paragraph{Proposition 2 (IR).}
For every advertiser $i$, nonnegative true type
$\mathbf v_i$, and profile of other reports
$\mathbf b_{-i}$, truthful participation yields
nonnegative utility:
\begin{equation}
    q_{i,k^{\mathrm T}}v_{i,k^{\mathrm T}}
    -p_i(\mathbf v_i,\mathbf b_{-i})
    \geq0,
    \qquad
    k^{\mathrm T}=k_\phi(\mathbf v_i,\mathbf b_{-i}).
    \label{eq:app-ir}
\end{equation}

\paragraph{Proof.}
Fix $i$, $\mathbf v_i$, and $\mathbf b_{-i}$.
Since $q_{ik}\geq0$, $v_{ik}\geq0$, and $w_i>0$,
for every configuration $k$ we have
\[
    S_\phi(k;\mathbf v_i,\mathbf b_{-i})
    =
    C_{ik}(\mathbf b_{-i})+w_iq_{ik}v_{ik}
    \geq C_{ik}(\mathbf b_{-i}).
\]
Taking maxima gives
\begin{equation}
    \max_{k\in\mathcal K}
    S_\phi(k;\mathbf v_i,\mathbf b_{-i})
    \geq
    \max_{k\in\mathcal K}C_{ik}(\mathbf b_{-i})
    =
    H_i(\mathbf b_{-i}).
    \label{eq:app-ama-ir-bound}
\end{equation}
Using the selected configuration's optimality and \eqref{eq:app-ama-utility-alignment}, truthful
utility is
\begin{align*}
    q_{i,k^{\mathrm T}}v_{i,k^{\mathrm T}}
    -p_i(\mathbf v_i,\mathbf b_{-i})
    &=
    \frac{
        S_\phi(k^{\mathrm T};\mathbf v_i,\mathbf b_{-i})
        -H_i(\mathbf b_{-i})
    }{w_i}
    \\
    &=
    \frac{
        \max_{k\in\mathcal K}
        S_\phi(k;\mathbf v_i,\mathbf b_{-i})
        -H_i(\mathbf b_{-i})
    }{w_i}
    \geq0.
\end{align*}
This proves \eqref{eq:app-ir}.

Equivalently, any
$\ell_i\in\arg\max_{\ell\in\mathcal K}
C_{i\ell}(\mathbf b_{-i})$
has menu price $P_{i\ell_i}(\mathbf b_{-i})=0$
and nonnegative true value.
Truthful reporting maximizes utility over all entries
and therefore performs at least as well as this entry.
The zero-priced entry may differ across advertisers;
it need not be configuration $0$.

\subsection{Nonnegative Payments and No Transaction}
\label{app:ama-payments}

\paragraph{Proposition 3.}
All menu prices and realized payments are nonnegative.
If configuration $0$ is selected, every advertiser's
realized payment is zero.

\paragraph{Proof.}
By definition,
$H_i(\mathbf b_{-i})\geq C_{ik}(\mathbf b_{-i})$
for every $k$.
Together with $w_i>0$, this implies
\[
    P_{ik}(\mathbf b_{-i})
    =
    \frac{H_i(\mathbf b_{-i})-C_{ik}(\mathbf b_{-i})}{w_i}
    \geq0.
\]
Hence the selected-entry payment $p_i(\mathbf b)$
is nonnegative.

Now suppose $k_\phi(\mathbf b)=0$.
Because $\beta_0=0$ and $q_{i0}=0$ for every $i$,
we have $S_\phi(0;\mathbf b)=0$.
Optimality of configuration $0$ therefore gives
$S_\phi(k;\mathbf b)\leq0$ for all $k$.
For every advertiser $i$ and configuration $k$,
nonnegative reports and CTI imply
\[
    C_{ik}(\mathbf b_{-i})
    =
    S_\phi(k;\mathbf b)-w_iq_{ik}b_{ik}
    \leq0.
\]
Since $C_{i0}(\mathbf b_{-i})=0$, we obtain
$H_i(\mathbf b_{-i})=0$.
Thus
\[
    p_i(\mathbf b)
    =
    P_{i0}(\mathbf b_{-i})
    =
    \frac{H_i(\mathbf b_{-i})-C_{i0}(\mathbf b_{-i})}{w_i}
    =0.
\]

For a general profile of other reports, however,
the no-allocation entry has menu price
$P_{i0}(\mathbf b_{-i})=H_i(\mathbf b_{-i})/w_i$,
which need not be zero when that entry is not selected.
The IR proof does not require configuration $0$
to have zero price in every menu.

\paragraph{Learning and deployment.}
The guarantees hold for every fixed parameter setting
with strictly positive advertiser weights.
They do not require revenue training to reach an optimum,
independence of private types, or monotonicity of CTI
in the target exposure shares.
The soft selection used during training is an
optimization surrogate.
The guarantees concern the deployed hard-max allocation
and the associated payments in \eqref{eq:ama-payment}.

\paragraph{Valuation-model scope.}
The guarantees are conditional on the valuation model
$V_i(k)=q_{ik}v_{ik}$ and apply pointwise in private types
and other advertisers' reports.
If $V_i(k)$ equals advertiser $i$'s true expected value
of the image generated under configuration $k$, the
results imply DSIC and IR in expectation over image
generation.
A fixed historical CTI estimate alone does not establish
this equality.
The results do not guarantee nonnegative realized
utility for every generated image or identify the
CTI proxy with an observed click-through rate.


\section{Theoretical Basis of Placement-Specific Evaluation Dimensions}
\label{app:cti-provement}
This section provides the theoretical basis for our four evaluation
dimensions. We first define the scope of factors considered in our framework,
then identify two complementary evaluation perspectives within this scope and
further divide each perspective into two categories, forming a $2\times2$
conceptual space. The four evaluation metrics are then positioned as the
corresponding operationalizations of the four conceptual cells.


\paragraph{Research scope.}
We focus on evaluation factors tailored to AIGC-based product placement rather
than directly adopting criteria from conventional advertising evaluation.
Specifically, we consider factors that are relevant to embedded product advertising and can be reliably assessed from generated images. Highly
subjective or abstract judgments, such as overall aesthetics or generic
attractiveness, are excluded because they are difficult to define consistently
and to evaluate robustly with automated models.


\paragraph{Two-dimensional conceptual space.}
We adopt the logic of \emph{conceptual typology}, which organizes a complex
construct through multiple dimensions and their combinations into a structured
property space~\citep{collier2012putting}. Following this methodology, we
characterize placement-specific influencing factors from two complementary
perspectives: \emph{Stimulus} and \emph{Processing}. This choice is grounded in
classical advertising information-processing theory, which distinguishes the
advertising stimuli or executional cues presented to consumers from the
subsequent processing through which these stimuli influence consumer
responses~\citep{macinnis1989information,macinnis1991enhancing}.

Formally, we define
\begin{equation}
    \mathcal{D}
    =
    \mathcal{S}
    \times
    \mathcal{P},
\end{equation}
where $\mathcal{S}$ and $\mathcal{P}$ denote the Stimulus and Processing
dimensions, respectively.


\paragraph{Stimulus: Product vs.\ Product--Scene Relation.}
In our framework, \emph{Stimulus} refers to the visual advertising content
from which a placement-specific influence originates. We divide this dimension
into the inserted product itself and the relation between the product and its
surrounding scene. This distinction is supported by established structural and
advertising theories: the Entity--Relationship model treats entities and their
relationships as different primitives~\citep{chen1976entity}, scene-graph
representations similarly separate objects, object attributes, and inter-object
relationships in visual scenes~\citep{johnson2015image,krishna2017visual}, and
product-placement research further shows that how a brand is presented and how
it is connected to host content can have different effects on brand memory and
attitude~\citep{russell2002investigating}. Together, these findings support
treating product properties and product--scene relational properties as two
distinct sources of placement-specific visual influence.

Accordingly, we define
\begin{equation}
    \mathcal{S}
    =
    \left\{
    \mathrm{Product},
    \mathrm{Product\text{-}Scene\ Relation}
    \right\}.
\end{equation}
Other visual elements are outside our research scope unless they interact with
the inserted product; such effects are subsumed under the
\emph{Product--Scene Relation} category.


\paragraph{Processing: Explicit vs.\ Implicit.}
In our framework, \emph{Processing} describes how visual advertising content
is interpreted by consumers. We distinguish between \emph{Explicit} processing,
which relies primarily on directly observable visual configurations such as
product size, position, and color, and \emph{Implicit} processing, which requires
further interpretation to derive higher-level commercial meanings such as brand
identity, product functionality, and perceived value. 

Accordingly, we define
\begin{equation}
    \mathcal{P}
    =
    \left\{
    \mathrm{Explicit},
    \mathrm{Implicit}
    \right\}.
\end{equation}

This distinction is
supported by advertising and consumer-processing theories: \cite{scott1994images}
characterizes advertising images as symbolic systems capable of conveying
meanings beyond their literal appearance; visual-rhetoric research further shows
that consumers construct higher-level meanings from visual configurations
~\citep{mcquarrie1999visual}; and means--end theory describes a progression from
concrete product attributes to increasingly abstract consequences and values
~\citep{reynolds1995means}.


\paragraph{Four conceptual cells.}
Crossing the two dimensions yields four conceptual cells, each corresponding
to a distinct source and mode of placement-specific influence. We instantiate
these cells with four evaluation dimensions, as summarized in
Table~\ref{tab:placement-conceptual-space}.

\begin{table}[h]
    \centering
    \small
    \setlength{\tabcolsep}{8pt}
    \renewcommand{\arraystretch}{1.12}
    \caption{The $2\times2$ conceptual organization of placement-specific
    visual evaluation.}
    \label{tab:placement-conceptual-space}
    \begin{tabular}{lcc}
        \toprule
        \textbf{Stimulus}
        & \multicolumn{2}{c}{\textbf{Processing}} \\
        \cmidrule(lr){2-3}
        & \textbf{Explicit}
        & \textbf{Implicit} \\
        \midrule
        Product
        & Product Prominence
        & Brand Recognizability \\
        Product--Scene Relation
        & Scene Integration
        & Contextual Value Communication \\
        \bottomrule
    \end{tabular}
\end{table}

\emph{Product Prominence} corresponds to the
\emph{Product$\times$Explicit} cell because it measures the directly observable
visual presence of the inserted product, reflected by factors such as its size,
position, and color.

\emph{Brand Recognizability} represents the
\emph{Product$\times$Implicit} cell. Although brand cues are physically present
on the product, their commercial meaning emerges only when visual features such
as logos, text, colors, or distinctive product designs are interpreted as a
particular brand identity.

\emph{Scene Integration} corresponds to the
\emph{Product--Scene Relation$\times$Explicit} cell because it evaluates
directly observable compatibility between the product and its surrounding
scene, including perspective, scale, illumination, contact, and occlusion.

\emph{Contextual Value Communication} represents the
\emph{Product--Scene Relation$\times$Implicit} cell. It captures whether the
relationship between the product and its context conveys higher-level
commercial meanings, such as product usage, functional benefits, lifestyle associations, or perceived value.

%
%
%
%

\paragraph{Scope of the completeness claim.}
We do not claim that the four metrics constitute all determinants of
click-through intent, nor that they exhaust every possible property within
their corresponding cells. Rather, our completeness claim concerns the
placement-specific conceptual space defined above. Within this space, many
plausible alternative metrics substantially overlap with the four selected
constructs, while others fall outside our scope because they are either not
particularly salient to AIGC product placement or cannot be evaluated reliably
from generated images by automated models. The four metrics therefore provide compact and operationalizable coverage of the major conceptual aspects
relevant to our setting.

\paragraph{Case-based analysis of alternative evaluation criteria.}
To assess the conceptual completeness of the four metrics within
the scope of this study, we examine whether any candidate criterion
satisfies both of the following conditions:
(i) it falls within our scope of evaluating product placement
in generated images; and
(ii) its primary evaluation content or focus differs substantially
from those of the four existing metrics, warranting its inclusion
as a separate dimension.
To this end, we analyze representative candidate criteria proposed
by advertising experts, together with common behavioral and
ad-delivery metrics.
For candidates within the research scope, we examine whether they
introduce distinct evaluation content or substantially overlap
conceptually with the existing metrics.
Table~\ref{tab:alternative-metrics} summarizes the results.
For brevity, PP, BR, SI, and CVC denote Product Prominence,
Brand Recognizability, Scene Integration, and Contextual Value
Communication, respectively.

The analysis indicates that, among the candidates examined,
those within the research scope substantially overlap conceptually
with the four existing metrics. The remaining candidates fall outside this scope. For example, some cannot be reliably assessed from image content alone, whereas others are conventional advertising metrics suited to evaluating carefully designed advertisements but not product placements in AI-generated images, where the primary objective is image generation rather than advertisement design.
These findings support the adequacy of the four proposed dimensions
in capturing the principal aspects of product-placement evaluation
within the scope of this study.

\begin{longtable}{
    C{0.18\linewidth}
    L{0.56\linewidth}
    C{0.16\linewidth}
}
\caption{Selected alternative evaluation criteria and their
relationships with the proposed metrics.}
\label{tab:alternative-metrics}\\

\toprule
\textbf{Candidate criterion}
&
\textbf{Case-based analysis}
&
\textbf{Relation to framework}
\\
\midrule
\endfirsthead

\toprule
\textbf{Candidate criterion}
&
\textbf{Case-based analysis}
&
\textbf{Relation to framework}
\\
\midrule
\endhead

Attention Retention
&
Sustained attention is a temporal behavioral response that cannot be measured directly from the image alone.
&
Out of scope
\\
\\
Product Recognizability
&
Identifying a product category or specific item is related to
identifying its brand.
&
BR
\\
\\
Viewpoint Favorability
&
Viewpoint affects product visibility, the visibility of brand cues,
and the plausibility of the product's presentation in the scene.
&
PP + BR + SI
\\
\\
Brand--Scene Congruence
&
Visual compatibility is addressed by SI, while the relevance
of the depicted context to product use or commercial meaning
is addressed by CVC.
&
SI + CVC
\\
\\
Brand--Plot Connection
&
In static images, the observable counterpart of plot connection
is the product's meaningful participation in the depicted situation,
which is addressed by CVC.
&
CVC
\\
\\
Placement Intrusiveness
&
Perceived intrusiveness is subjective but often arises from unnatural product placement.
&
SI
\\
\\
Ad--Product Relevance
&
Whether the surrounding content supports the product's use,
benefits, or positioning falls within the contextual meaning
evaluated by CVC.
&
CVC
\\
\\
Informativeness
&
Informativeness is not necessarily associated with a higher click-through rate.
&
Out of scope
\\
\\
Processing Fluency
&
Processing fluency is a broad concept influenced by multiple factors and is difficult to assess directly from an image alone. Product prominence, recognizable brand cues, and clear communication of contextual product value may facilitate processing. 
&
PP + BR + CVC
\\
\\
Attitude toward the Brand
&
Brand attitude reflects consumer evaluations that may predate
the image and vary with experience and preference. Such
consumer-specific information is outside our evaluation scope.
&
Out of scope
\\
\\
Brand Recall / Memorability
&
Retention and retrieval of brand information require
post-exposure assessment. 
&
Out of scope
\\
\\
Overall Aesthetics
&
Overall aesthetics concerns the composition, style, and artistic quality of the entire image. However, in image advertising, we focus more on advertising effectiveness than on aesthetic appeal.
&
Out of scope
\\
\\
Creativity / Novelty
&
Creativity and novelty concern the originality of a placement relative to prior examples and expectations. Although they may be related to CTI, our setting involves naturally integrating products into scenes rather than creating carefully designed image advertisements. Creativity is therefore generally not a central consideration in our setting.
&
Out of scope
\\
\\
Persuasiveness
&
Persuasiveness concerns changes in consumer beliefs, attitudes,
or intentions. It depends on message interpretation and consumer
characteristics beyond the placement properties assessed here.
&
Out of scope
\\
\\
Purchase Intention
&
Purchase intention depends on product need, price, trust,
and purchasing constraints in addition to visual presentation.
These factors extend beyond our content-based evaluation.
&
Out of scope
\\
\\
User--Product Relevance
&
Relevance to a particular user depends on that user's interests,
needs, and preferences. It requires audience information
beyond the depicted product--scene relationship.
&
Out of scope
\\

\bottomrule
\end{longtable}

\section{Prompts for CTI evaluation}
\label{app:cti}

In this section, we provide the prompts used for CTI prediction, including the prompt for the VLM to extract visual evidence from images and the prompt for the LLM to score the extracted visual evidence.
\subsection{Visual Evidence Extraction Prompt}
\label{app:visual-evidence-prompt}

\begin{promptbox}{Visual-evidence extraction prompt}
You are a visual analysis expert. Examine the input image and extract visual evidence for each target product for subsequent model scoring.

Scene/context:
{base_prompt}

Target products:
product_A: {agent1_prompt}
product_B: {agent2_prompt}

General Requirements:

- If a target product is absent, state "not visible." If it cannot be reliably matched to an object in the image, state "not clearly visible" or "uncertain." Do not infer its brand, location, or relationship to the scene.

- Distinguish between product visibility and confirmation of the target brand. Describe all logos, text, and patterns that are actually visible. If text is illegible, explicitly state so rather than reconstructing or completing it.

- If an object that may correspond to a target product displays branding inconsistent with the specified target brand, explicitly describe the visible branding and the mismatch. Do not attribute branding found on other objects to the target product.

- Describe visible facts first, followed only by limited interpretations supported by those facts. Do not replace concrete evidence with vague conclusions such as "highly congruent," "highly recognizable," or "naturally integrated."

Extract evidence for each product along the following four dimensions:

1. Product Prominence

Describe the product's position, depth within the scene, visible area, occlusion or cropping, brightness, clarity, and other relevant visual properties.

Compare the two target products in terms of visible area, proportion of the image occupied, and overall visual prominence. If a comparison cannot be made, explain why. Do not equate visibility with prominence, infer prominence from size alone, or fabricate precise proportions.

2. Brand Recognizability

Record any visible logos, text, patterns, or distinctive design features. Explain whether these cues support identification of the target brand or only recognition of the general product category. Do not confirm a brand solely from the target product name, generic colors, or ordinary packaging.

3. Scene Integration

Describe the product's contact with and support from its surroundings, as well as its scale, perspective, lighting, shadows, boundaries, and occlusion relationships.

Identify any visible floating, interpenetration, ghosting, pasted-on appearance, or structural abnormalities. If these properties cannot be assessed, state that they are uncertain. Do not use suitability for the usage context as a substitute for physical and visual evidence.

4. Contextual Value Communication

Describe whether the scene represents a typical usage context for the product, whether it demonstrates the product's use, functional benefits, or lifestyle associations, and whether the visible context supports the product's promotional communication.

Output Requirements:

- Output valid JSON only.

- If a target product is not visible or cannot be reliably identified, state this explicitly in each relevant field without fabricating information.

- Do not provide scores, ratings, or conclusions about overall advertising effectiveness. Report only visually grounded evidence.

Output Format:

{
  "product_A": {
    "product_prominence_evidence": "",
    "brand_recognizability_evidence": "",
    "scene_integration_evidence": "",
    "contextual_value_communication_evidence": ""
  },
  "product_B": {
    "product_prominence_evidence": "",
    "brand_recognizability_evidence": "",
    "scene_integration_evidence": "",
    "contextual_value_communication_evidence": ""
  }
}
\end{promptbox}

\subsection{CTI Scoring Prompt}
\label{app:cti-scoring-prompt}

\begin{promptbox}{CTI scoring prompt}
You are a professional advertising effectiveness evaluator.

Below is objective visual evidence extracted by a VLM from a single advertisement image. Based only on the provided evidence, assign each product an integer score from 1 to 5 on each of the following four dimensions, where 5 indicates the best performance and 1 indicates the worst.

1. Product Prominence
2. Brand Recognizability
3. Scene Integration
4. Contextual Value Communication

Important:

- This is a single-image evaluation task. Do not refer to or compare with other images.
- Base your ratings only on the provided VLM visual evidence. Do not introduce information beyond the evidence.
- Evaluate each dimension independently.
- If the target product itself is not visible, do not assign a high score merely because the scene contains products from the same category, similar products, logos of other brands, or textual mentions of the product.
- Do not output explanations, Markdown, code blocks, or your reasoning process.

Scene/context:
{Current scene description}

Products to evaluate:
product_A: {Product A description}
product_B: {Product B description}

VLM visual evidence:
{
  "product_A": {
    "product_prominence_evidence": "{VLM-extracted prominence evidence for Product A}",
    "brand_recognizability_evidence": "{VLM-extracted brand recognizability evidence for Product A}",
    "scene_integration_evidence": "{VLM-extracted scene integration evidence for Product A}",
    "contextual_value_communication_evidence": "{VLM-extracted contextual value communication evidence for Product A}"
  },
  "product_B": {
    "product_prominence_evidence": "{VLM-extracted prominence evidence for Product B}",
    "brand_recognizability_evidence": "{VLM-extracted brand recognizability evidence for Product B}",
    "scene_integration_evidence": "{VLM-extracted scene integration evidence for Product B}",
    "contextual_value_communication_evidence": "{VLM-extracted contextual value communication evidence for Product B}"
  }
}

Scoring criteria:

1. Product Prominence

5. Highly prominent
- The product occupies a large portion of the image and is positioned near the visual center or another highly salient location.
- The product exhibits strong contrast with the background in color, brightness, texture, or spatial arrangement.
- The product strongly attracts visual attention and serves as a primary focus of the image.

4. Relatively prominent
- The product has a clear visual presence, occupying a relatively large portion of the image or appearing in a prominent position.
- The product is unobstructed or only slightly occluded.
- The product is easy to notice but is not necessarily the dominant focal point of the image.

3. Moderately prominent
- The product is clearly visible, with a moderate size and ordinary placement.
- The product has a noticeable visual presence but no clear visual advantage over other elements in the image.

2. Low prominence
- The product is relatively small, positioned away from the visual center, or noticeably occluded.
- Other visual elements may distract attention from the product, making it unlikely to be noticed immediately.

1. Very low prominence
- The product occupies only a very small portion of the image, is severely occluded, or is barely visible.
- The product is difficult for viewers to notice.

2. Brand Recognizability

5. Highly recognizable
- The logo, brand name, packaging, or other distinctive brand cues are clear and prominent.
- Brand information is unobstructed and contains no obvious generation errors.
- Viewers can quickly and accurately identify the brand.

4. Relatively recognizable
- Clear logos, brand names, packaging, or other brand cues are visible.
- Viewers can identify the specific brand relatively easily, although the brand cues are less prominent than at score 5.

3. Moderately recognizable
- Recognizable brand cues are present but may be small, positioned away from the visual center, or partially occluded.
- Viewers can still identify the specific brand from the available visual information.

2. Low recognizability
- Brand cues are blurred, heavily occluded, too small, or affected by noticeable generation errors.
- The product category may be recognizable, but the specific brand is difficult to identify reliably.

1. Unrecognizable
- Reliable logos, brand names, packaging, or other distinctive brand cues are absent.
- Viewers cannot determine the specific brand based solely on the visual information.

3. Scene Integration

5. Highly integrated
- The product is highly consistent with the scene in terms of spatial placement, scale, perspective, lighting, shadows, and boundaries.
- Spatial relationships between the product and surrounding objects are natural and physically plausible.
- There are no obvious generation artifacts, and the overall appearance resembles a real photograph.

4. Well integrated
- The product is naturally integrated into the scene overall.
- Spatial relationships, scale, perspective, and lighting are generally plausible, with only minor imperfections.

3. Moderately integrated
- The product is reasonably integrated into the scene but exhibits some visible generation artifacts.
- Minor inconsistencies may appear in lighting, boundaries, scale, perspective, or local details.

2. Poorly integrated
- The product is noticeably inconsistent with its surrounding environment.
- There may be obvious compositing artifacts, perspective errors, abnormal scale, inconsistent lighting, unnatural boundaries, or implausible spatial relationships.

1. Not integrated
- The product clearly fails to integrate naturally into the scene.
- Severe artifacts are present, such as floating objects, ghosting, structural abnormalities, major scale or perspective errors, or an obvious pasted-on appearance.

4. Contextual Value Communication

5. Highly effective
- The scene clearly demonstrates or reinforces the product's use, usage scenario, relevance, or core value.
- The product forms clear and meaningful relationships with surrounding people, objects, or environmental elements.
- Viewers can quickly understand from the scene why the product is useful or appealing.

4. Relatively effective
- The scene is strongly related to the product's use or value.
- The scene helps viewers understand the product's usage context or value, although the communication is weaker than at score 5.

3. Moderately effective
- The scene does not conflict with the product and provides some context regarding its use or value.
- However, the scene provides only limited reinforcement of the product's use, relevance, or value.

2. Low effectiveness
- The scene has only a weak relationship with the product's use or value.
- Although the product appears in the scene, the surrounding context provides little help in understanding its use, relevance, or value.

1. No effective communication
- The scene fails to communicate the product's use, relevance, or value, or clearly conflicts with the intended use conveyed by the product.
- Viewers cannot obtain meaningful information about the product's value from the scene.

Return only valid JSON. Replace each <score> with an integer from 1 to 5. Do not output any additional content:

{
  "product_A": {
    "product_prominence": <score>,
    "brand_recognizability": <score>,
    "scene_integration": <score>,
    "contextual_value_communication": <score>
  },
  "product_B": {
    "product_prominence": <score>,
    "brand_recognizability": <score>,
    "scene_integration": <score>,
    "contextual_value_communication": <score>
  }
}
\end{promptbox}

\end{document}

%% file: math_commands.tex
\usepackage{amsmath,amsfonts,bm}

\def\eqref#1{equation~\ref{#1}}

\def\1{\bm{1}}

\def\mR{{\bm{R}}}

\def\mU{{\bm{U}}}

\def\mW{{\bm{W}}}

\def\mY{{\bm{Y}}}

\DeclareMathAlphabet{\mathsfit}{\encodingdefault}{\sfdefault}{m}{sl}
\SetMathAlphabet{\mathsfit}{bold}{\encodingdefault}{\sfdefault}{bx}{n}

\def\sT{{\mathbb{T}}}

\newcommand{\R}{\mathbb{R}}



%% file: references.bib
@article{collier2012putting,
  title={Putting typologies to work: Concept formation, measurement, and analytic rigor},
  author={Collier, David and LaPorte, Jody and Seawright, Jason},
  journal={Political research quarterly},
  volume={65},
  number={1},
  pages={217--232},
  year={2012},
  publisher={Sage Publications Sage CA: Los Angeles, CA}
}

@article{macinnis1989information,
  title={Information processing from advertisements: Toward an integrative framework},
  author={MacInnis, Deborah J and Jaworski, Bernard J},
  journal={Journal of marketing},
  volume={53},
  number={4},
  pages={1--23},
  year={1989},
  publisher={SAGE Publications Sage CA: Los Angeles, CA}
}

@article{macinnis1991enhancing,
  title={Enhancing and Measuring Consumers' Motivation, Opportunity, and Ability to Process Brand Information from Ads},
  author={MacInnis, Deborah J. and Moorman, Christine and Jaworski, Bernard J.},
  journal={Journal of Marketing},
  volume={55},
  number={4},
  pages={32--53},
  year={1991}
}

@article{scott1994images,
  title={Images in Advertising: The Need for a Theory of Visual Rhetoric},
  author={Scott, Linda M.},
  journal={Journal of Consumer Research},
  volume={21},
  number={2},
  pages={252--273},
  year={1994}
}

@article{mcquarrie1999visual,
  title={Visual Rhetoric in Advertising: Text-Interpretive, Experimental, and Reader-Response Analyses},
  author={McQuarrie, Edward F. and Mick, David Glen},
  journal={Journal of Consumer Research},
  volume={26},
  number={1},
  pages={37--54},
  year={1999}
}

@article{reynolds1995means,
  title={A Means-End Analysis of Brand Persuasion Through Advertising},
  author={Reynolds, Thomas J. and Gengler, Charles E. and Howard, Daniel J.},
  journal={International Journal of Research in Marketing},
  volume={12},
  number={3},
  pages={257--266},
  year={1995}
}

@article{chen1976entity,
  title={The Entity-Relationship Model---Toward a Unified View of Data},
  author={Chen, Peter Pin-Shan},
  journal={ACM Transactions on Database Systems},
  volume={1},
  number={1},
  pages={9--36},
  year={1976}
}

@inproceedings{johnson2015image,
  title={Image Retrieval Using Scene Graphs},
  author={Johnson, Justin and Krishna, Ranjay and Stark, Michael and
          Li, Li-Jia and Shamma, David A. and Bernstein, Michael S.
          and Fei-Fei, Li},
  booktitle={Proceedings of the IEEE Conference on Computer Vision
             and Pattern Recognition},
  pages={3668--3678},
  year={2015}
}

@article{russell2002investigating,
  title={Investigating the Effectiveness of Product Placements in Television
         Shows: The Role of Modality and Plot Connection Congruence on
         Brand Memory and Attitude},
  author={Russell, Cristel Antonia},
  journal={Journal of Consumer Research},
  volume={29},
  number={3},
  pages={306--318},
  year={2002}
}

@article{edelman2007internet,
  title={Internet advertising and the generalized second-price auction: Selling billions of dollars worth of keywords},
  author={Edelman, Benjamin and Ostrovsky, Michael and Schwarz, Michael},
  journal={American economic review},
  volume={97},
  number={1},
  pages={242--259},
  year={2007},
  publisher={American Economic Association}
}

@article{varian2007position,
  title={Position auctions},
  author={Varian, Hal R},
  journal={international Journal of industrial Organization},
  volume={25},
  number={6},
  pages={1163--1178},
  year={2007},
  publisher={Elsevier}
}

@inproceedings{duetting2024mechanism,
  title={Mechanism design for large language models},
  author={Duetting, Paul and Mirrokni, Vahab and Paes Leme, Renato and Xu, Haifeng and Zuo, Song},
  booktitle={Proceedings of the ACM Web Conference 2024},
  pages={144--155},
  year={2024}
}

@article{hajiaghayi2024ad,
  title={Ad auctions for llms via retrieval augmented generation},
  author={Hajiaghayi, MohammadTaghi and Lahaie, S{\'e}bastien and Rezaei, Keivan and Shin, Suho},
  journal={Advances in Neural Information Processing Systems},
  volume={37},
  pages={18445--18480},
  year={2024}
}

@inproceedings{sun2025bidding,
  title     = {Bidding for Influence: Auction-Driven Diffusion Image Generation},
  author    = {Sun, Lillian and Huang, Henry and Zhu, Fucheng Warren and
               Daras, Giannis and Daskalakis, Constantinos Costis},
  booktitle = {ICML 2025 Workshop on Collaborative and Federated Agentic Workflows},
  year      = {2025},
  url       = {https://openreview.net/forum?id=v5tF1THvt9}
}

@inproceedings{soumalias2026truthful,
 author = {Soumalias, Ermis and Curry, Michael and Seuken, Sven},
 booktitle = {Advances in Neural Information Processing Systems},
 doi = {10.52202/085713-4862},
 editor = {D. Belgrave and C. Zhang and H. Lin and R. Pascanu and P. Koniusz and M. Ghassemi and N. Chen},
 pages = {145262--145308},
 publisher = {Curran Associates, Inc.},
 title = {Truthful Aggregation of LLMs with an Application to Online Advertising},
 url = {https://proceedings.neurips.cc/paper_files/paper/2025/file/d60b6b7f0ba6bf07d975b3bbdacea702-Paper-Conference.pdf},
 volume = {38, Main Conference},
 year = {2025}
}

@article{zhao2025llm,
  title={Llm-auction: Generative auction towards llm-native advertising},
  author={Zhao, Chujie and Hu, Qun and Song, Shiping and Chen, Dagui and Zhu, Han and Xu, Jian and Zheng, Bo},
  journal={arXiv preprint arXiv:2512.10551},
  year={2025}
}

@article{yun2026llm,
  title={LLM Advertisement based on Neuron Auctions},
  author={Yun, Peiran and Xu, Wenxin and Liu, Jiayuan and Zhang, Yihang and Zeng, Liang and Kong, Lingkai and Wang, Tonghan},
  journal={arXiv preprint arXiv:2605.08326},
  year={2026}
}

@article{myerson1981optimal,
  title={Optimal auction design},
  author={Myerson, Roger B},
  journal={Mathematics of operations research},
  volume={6},
  number={1},
  pages={58--73},
  year={1981},
  publisher={INFORMS}
}

@article{hertz2022prompt,
  title={Prompt-to-prompt image editing with cross attention control},
  author={Hertz, Amir and Mokady, Ron and Tenenbaum, Jay and Aberman, Kfir and Pritch, Yael and Cohen-Or, Daniel},
  journal={arXiv preprint arXiv:2208.01626},
  year={2022}
}

@inproceedings{ge2023expressive,
  title={Expressive text-to-image generation with rich text},
  author={Ge, Songwei and Park, Taesung and Zhu, Jun-Yan and Huang, Jia-Bin},
  booktitle={2023 IEEE/CVF International Conference on Computer Vision (ICCV)},
  pages={7511--7522},
  year={2023},
  organization={IEEE}
}

@article{brack2023sega,
  title={Sega: Instructing text-to-image models using semantic guidance},
  author={Brack, Manuel and Friedrich, Felix and Hintersdorf, Dominik and Struppek, Lukas and Schramowski, Patrick and Kersting, Kristian},
  journal={Advances in Neural Information Processing Systems},
  volume={36},
  pages={25365--25389},
  year={2023}
}

@inproceedings{dalva2025fluxspace,
  title={Fluxspace: Disentangled semantic editing in rectified flow models},
  author={Dalva, Yusuf and Venkatesh, Kavana and Yanardag, Pinar},
  booktitle={2025 IEEE/CVF Conference on Computer Vision and Pattern Recognition (CVPR)},
  pages={13083--13092},
  year={2025},
  organization={IEEE}
}

@inproceedings{curry2023differentiable,
  title={Differentiable Economics for Randomized Affine Maximizer Auctions},
  author={Curry, Michael and Sandholm, Tuomas and Dickerson, John},
  booktitle={Proceedings of the Thirty-Second International Joint Conference on Artificial Intelligence},
  pages={2633--2641},
  year={2023},
  doi={10.24963/ijcai.2023/293}
}

@misc{turner2026evaluating,
  title={Evaluating and Pricing Advertisements in AI-Generated Responses},
  author={Turner-Smith, John L. and Huang, Zimeng and Fu, Yuhan and Zhang, Yihang and Wang, Tonghan},
  year={2026},
  eprint={2607.27686},
  archivePrefix={arXiv},
  primaryClass={cs.AI}
}

@article{zhang2026pila,
  title={PILA: Plug-and-Play Insertion for LLM-native Advertising},
  author={Zhang, Zhaowei and Fu, Yuhan and Zhang, Yihang and Liu, Xiaohan and Zhang, Ceyao and Zhang, Xiaoyuan and Kang, Yipeng and Wang, Tonghan and Yang, Yaodong},
  journal={arXiv preprint arXiv:2607.25590},
  year={2026}
}

@article{zhang2026naiad,
  title={NaiAD: Initiate Data-Driven Research for LLM Advertising},
  author={Zhang, Yihang and Huang, Zimeng and Zhai, Ren and Kang, Yipeng and Wang, Tonghan},
  journal={arXiv preprint arXiv:2605.09918},
  year={2026}
}

@inproceedings{liu2026incentive,
  title={Incentive-Aware Multi-Fidelity Optimization for Generative Advertising in Large Language Models},
  author={Liu, Jiayuan and Wang, Barry and Gan, Jiarui and Wang, Tonghan and Xie, Leon and Guo, Mingyu and Conitzer, Vincent},
  booktitle={Proceedings of the Third Conference on Language Modeling},
  year={2026}
}

@article{krishna2017visual,
  title   = {Visual Genome: Connecting Language and Vision Using
             Crowdsourced Dense Image Annotations},
  author  = {Krishna, Ranjay and Zhu, Yuke and Groth, Oliver and
             Johnson, Justin and Hata, Kenji and Kravitz, Joshua and
             Chen, Stephanie and Kalantidis, Yannis and Li, Li-Jia and
             Shamma, David A. and Bernstein, Michael S. and
             Fei-Fei, Li},
  journal = {International Journal of Computer Vision},
  volume  = {123},
  number  = {1},
  pages   = {32--73},
  year    = {2017},
  doi     = {10.1007/s11263-016-0981-7}
}

@article{zhu2025internvl3,
  title   = {InternVL3: Exploring Advanced Training and Test-Time Recipes
             for Open-Source Multimodal Models},
  author  = {Zhu, Jinguo and Wang, Weiyun and Chen, Zhe and Liu, Zhaoyang
             and Ye, Shenglong and Gu, Lixin and Tian, Hao and Duan, Yuchen
             and Su, Weijie and Shao, Jie and others},
  journal = {arXiv preprint arXiv:2504.10479},
  year    = {2025},
  url     = {https://arxiv.org/abs/2504.10479}
}

@article{yang2024qwen25,
  title   = {Qwen2.5 Technical Report},
  author  = {Yang, An and Yang, Baosong and Zhang, Beichen and Hui, Binyuan
             and Zheng, Bo and Yu, Bowen and Li, Chengyuan and Liu, Dayiheng
             and Huang, Fei and Wei, Haoran and others},
  journal = {arXiv preprint arXiv:2412.15115},
  year    = {2024},
  url     = {https://arxiv.org/abs/2412.15115}
}

@article{vickrey1961counterspeculation,
  title   = {Counterspeculation, Auctions, and Competitive Sealed Tenders},
  author  = {Vickrey, William},
  journal = {The Journal of Finance},
  volume  = {16},
  number  = {1},
  pages   = {8--37},
  year    = {1961},
  doi     = {10.1111/j.1540-6261.1961.tb02789.x}
}

@article{clarke1971multipart,
  title   = {Multipart Pricing of Public Goods},
  author  = {Clarke, Edward H.},
  journal = {Public Choice},
  volume  = {11},
  pages   = {17--33},
  year    = {1971},
  doi     = {10.1007/BF01726210}
}

@article{groves1973incentives,
  title   = {Incentives in Teams},
  author  = {Groves, Theodore},
  journal = {Econometrica},
  volume  = {41},
  number  = {4},
  pages   = {617--631},
  year    = {1973},
  doi     = {10.2307/1914085}
}

@inproceedings{wang2024gemnet,
  title={GemNet: Menu-Based, strategy-proof multi-bidder auctions through deep learning},
  author={Wang, Tonghan and Jiang, Yanchen and Parkes, David C},
  booktitle={Proceedings of the 25th ACM Conference on Economics and Computation},
  year={2024}
}

@inproceedings{wang2025bundleflow,
  title     = {BundleFlow: Deep Menus for Combinatorial Auctions by Diffusion-Based Optimization},
  author    = {Wang, Tonghan and Jiang, Yanchen and Parkes, David C.},
  booktitle = {Advances in Neural Information Processing Systems (NeurIPS)},
  year      = {2025},
}

@article{hammond1979straightforward,
  title={Straightforward individual incentive compatibility in large economies},
  author={Hammond, Peter J},
  journal={The Review of Economic Studies},
  volume={46},
  number={2},
  pages={263--282},
  year={1979},
  publisher={Wiley-Blackwell}
}

@article{schwartz2017customer,
  author  = {Schwartz, Eric M. and Bradlow, Eric T. and Fader, Peter S.},
  title   = {Customer Acquisition via Display Advertising Using Multi-Armed Bandit Experiments},
  journal = {Marketing Science},
  year    = {2017},
  volume  = {36},
  number  = {4},
  pages   = {500--522},
  doi     = {10.1287/mksc.2016.1023},
  url     = {https://doi.org/10.1287/mksc.2016.1023}
}

@article{li2025impact,
  author  = {Li, Yuan and Deng, Xiaoyu and Wu, Banggang},
  title   = {The Impact of Mobile Advertising Cue Types on Consumer Response Behaviors: Evidence from a Field Experiment},
  journal = {Journal of Theoretical and Applied Electronic Commerce Research},
  year    = {2025},
  volume  = {20},
  number  = {3},
  pages   = {244},
  doi     = {10.3390/jtaer20030244},
  url     = {https://doi.org/10.3390/jtaer20030244}
}
